\documentclass[]{article}

\usepackage{epstopdf}
\usepackage{subfigure}

\usepackage{natbib}
\usepackage{graphicx}
\usepackage{amsmath}
\usepackage{booktabs}
\bibpunct[, ]{(}{)}{;}{a}{}{,}
\renewcommand\bibfont{\fontsize{10}{12}\selectfont}

\usepackage{hyperref}
\usepackage[utf8]{inputenc}
\def\tightlist{}
\usepackage{float}
\restylefloat{table}

\begin{document}

\title{Transparency, Auditability and eXplainability of Machine Learning Models
in Credit Scoring}

\author{Michael Bücker\\FH Münster - University of Applied Sciences \\Münster School of Business \and Gero Szepannek \\ OST -- Stralsund University of Applied Sciences \\ Department of
Economics \and Alicja Gosiewska \\ Warsaw University of Technology \\ Faculty of Mathematics and Information
Science  \and Przemyslaw Biecek \\ University of Warsaw \\ Faculty of Mathematics, Informatics, and
Mechanics }

\thanks{CONTACT Michael Bücker. Email: \href{mailto:michael.buecker@fh-muenster.de}{\nolinkurl{michael.buecker@fh-muenster.de}}, Gero Szepannek. Email: \href{mailto:gero.szepannek@hochschule-stralsund.de}{\nolinkurl{gero.szepannek@hochschule-stralsund.de}}, Alicja Gosiewska. Email: \href{mailto:a.gosiewska@mini.pw.edu.pl}{\nolinkurl{a.gosiewska@mini.pw.edu.pl}}, Przemyslaw Biecek. Email: \href{mailto:p.biecek@mimuw.edu.pl}{\nolinkurl{p.biecek@mimuw.edu.pl}}}

\maketitle

\begin{abstract}
A major requirement for credit scoring models is to provide a maximally
accurate risk prediction. Additionally, regulators demand these models
to be transparent and auditable. Thus, in credit scoring, very simple
predictive models such as logistic regression or decision trees are
still widely used and the superior predictive power of modern machine
learning algorithms cannot be fully leveraged. Significant potential is
therefore missed, leading to higher reserves or more credit defaults.
This paper works out different dimensions that have to be considered for
making credit scoring models understandable and presents a framework for
making ``black box'' machine learning models transparent, auditable and
explainable. Following this framework, we present an overview of
techniques, demonstrate how they can be applied in credit scoring and
how results compare to the interpretability of score cards. A real world
case study shows that a comparable degree of interpretability can be
achieved while machine learning techniques keep their ability to improve
predictive power.
\end{abstract}

\hypertarget{introduction}{%
\section{Introduction}\label{introduction}}

The field of interpretable machine learning has rapidly advanced in
recent years. There are at least three reasons for this. First, there
are an increasing number of predictive models that affect our everyday
life. A popular example is credit decisions based on scoring models. The
unprecedented scale in which autonomous systems affect our lives bring
people's attention to a potential negative impact of such automation,
which has led to new regulations such as GDPR
\citep{Goodman_Flaxman_2017} and ehtical guidelines such as the one
published by the \citet{EthicsEU2019}.

Secondly, there is an increasing availability of large datasets and
cheap computational power. Traditionally, predictive models were built
mostly based on domain knowledge. Recent machine learning algorithms
automatically seek patterns in collected data, which is different from
the model developer perspective: Models based on domain knowledge are
understandable to model developers. Elastic and complex data-hungry
models are not necessarily transparent anymore.

For the application of machine learning in financial services the
Finacial Stability Board (FSB) states that \emph{``the use of complex
algorithms could result in a lack of transparency to consumers. This
`black box' aspect of machine learning algorithms may in turn raise
concerns. When using machine learning to assign credit scores and make
credit decisions, it is generally more difficult to provide consumers,
auditors, and supervisors with an explanation of a credit score and
resulting credit decision if challenged''} \citep{fsb2017}. Thus, model
developers are confronted with an increasing need for tools to better
understand what their models have learned.

Thirdly, there is a growing number of reported failures of complex
predictive models in different domains in the recent past that can be
traced back to a lack of proper model validation and model understanding
\citep{10.1371/journal.pone.0151470, nytimesJail, airQuality, ONeil}.

In the context of credit risk modelling there are several specific
requirements for models, which have led to consistent popularity of
logistic regression models \citep{sze2017ont}:

\begin{enumerate}
\def\labelenumi{\arabic{enumi}.}
\tightlist
\item
  the models are subject to regulation and auditors are typically
  familiar with interpreting logistic regression models due to their
  linear nature,
\item
  regulation further requires recurrent monitoring, which can be easily
  interpreted on a variable level for logistic regression models,
\item
  customers have a right to explanation of individual decisions, and an
  answer to the question why a credit application has been rejected can
  be easily broken down since the score computed from a logistic
  regression model is the sum of the variables' effects.
\end{enumerate}

Thus, based on the above reasons, standard logistic regression is
currently still considered as the gold standard methodology in the
credit scoring industry, despite the general superiority of modern
machine learning techniques
\citep[cf.~e.g.][]{baes02ben, les15ben, lou16clas, bis2014onc, sze2017ont, brown2012, belotti2009, fitz2016}.

For Machine Learning models which often act as black-boxes and lack the
transparency of simpler models such as logistic regression, many
techniques have been developed that help to create explanations for
models and predictions \citep[cf.][]{mol2019int}. However, a
standardized framework for these explainable ML (sometimes also denoted
as explainable AI, abbrev. XAI) methods that can provide structured
guidance on how to apply them in an application such as credit scoring
is still missing.

\hypertarget{framework-for-transparency-auditability-and-explainability-of-models-for-credit-scoring}{%
\section{Framework for Transparency, Auditability and eXplainability of
Models for Credit
Scoring}\label{framework-for-transparency-auditability-and-explainability-of-models-for-credit-scoring}}

\hypertarget{requirements-for-credit-scoring}{%
\subsection{Requirements for Credit
Scoring}\label{requirements-for-credit-scoring}}

There are several specific requirements to model exploration in the
context of credit scoring. In this section, these requirements will be
detaied out. In the following subsections we will map these requirements
to specific XAI methods.

An important set of requirements on the transparency of models are based
on the Basel Commitee on Banking Supervision (BCBS): According to the
European Banking Authority \emph{``the selection of certain risk drivers
and rating criteria should be based not only on statistical analysis,
but {[}that{]} the relevant business experts should be consulted on the
business rationale and risk contribution of the risk drivers under
consideration''} \citep{eba2017}. Companies should establish checks and
tests at each stage of the development process and be able to
\emph{``demonstrate developmental evidence of theoretical construction,
behavioural characteristics and key assumptions, types and use of input
data, numerical analysis routines and specified mathematical
calculations, and code writing language and protocols (to replicate the
model)''} \citep{fsb2017}. While the last two aspects refer to
reproducibility and proper documentation the first three ones are
closely related to global-level explanations as described in section
4.1.

One important aspect of data protection regulations is the right to
explanation. Specifically, the European Union General Data Protection
Regulation states, that a person whose personal data is used
\emph{``shall have the right not to be subject to a decision based
solely on automated processing, including profiling, which produces
legal effects concerning him or her or similarly significantly affects
him or her''} \citep[Article 22 (1)]{EUdataregulations2018}. Also,
\emph{``the data controller shall implement suitable measures to
safeguard the data subject's rights and freedoms and legitimate
interests, at least the right to obtain human intervention on the part
of the controller, to express his or her point of view and to contest
the decision''} \citep[Article 22 (3)]{EUdataregulations2018}. This
requires transparency of every single credit decision supported by data
and algorithms. Thus, it is not sufficient for banks to be able to
explain a credit scoring algorithm and how it is making predictions in
general but also to explain any single credit decision and to identify
the most \emph{adverse characteristics} of an applicant which means
being able to provide instance-level explanations as described in
section 4.2.

According to the Financial Stability Board the use of machine learning
techniques in combination with new data sources bears the potential of
risk assessment also for customers without known credit experience while
at the same time risking to introduce bias or ethical issues into
modelling, e.g.~by learning models that yield combinations of borrower
characteristics that are nothing but correlates of race or gender
\citep[Sec. 3.1.1 and Annex B]{fsb2017}. A road towards ethical fairness
of machine learning models is discussed in \citep{kusner2020}. As
outlined in \citep{garz2019} ethical considerations will become an
increasing challenge for the future role of data scientists. In April
2019, the European Commission High-Level Expert Group on AI presented
``Ethics Guidelines for Trustworthy Artificial Intelligence''
\citep{EthicsEU2019}. Three of the guidelines directly refer to
explainability:

\begin{enumerate}
\def\labelenumi{\arabic{enumi}.}
\tightlist
\item
  \textbf{Human agency and oversight:} proper oversight mechanisms need
  to be ensured, which can be achieved through human-in-the-loop
  approach
\item
  \textbf{Transparency:} AI systems and their decisions should be
  explained in a manner adapted to the stakeholder concerned and
\item
  \textbf{Accountability:} Mechanisms should be put in place to ensure
  responsibility and accountability for AI systems and their outcomes.
\end{enumerate}

These requirements imply that the problem of explainability should be
addressed more broadly than just from the perspective of the bank's data
science division, or from the perspective of the customer. The topic of
explainability should be analysed from the perspective of all
stakeholders present in the in the model life cycle
\citep{arya2019explanation, Sokol2020}. The topic of explainability and
ethics of AI appears in EU documents not only in the context of
requirements protecting civil rigths, but also in the context of
strategies towards responsible algorithms \citep{Excellence20}.

\hypertarget{proposed-framework-for-transparency-auditability-and-explainability}{%
\subsection{Proposed Framework for Transparency, Auditability and
eXplainability}\label{proposed-framework-for-transparency-auditability-and-explainability}}

In this section we introduce a systematic model exploration process
focused on \textbf{T}ransparency, \textbf{A}uditability and
e\textbf{X}plainability for \textbf{C}redit \textbf{S}coring models
(TAX4CS). Its outline is presented in the Figure \ref{fig:xaiiceberg}.

\hypertarget{stakeholders}{%
\subsubsection{Stakeholders}\label{stakeholders}}

As recommended in \citep{EthicsEU2019}, the first step of the process is
to identify the stakeholders that participate in the model life cycle.
These stakeholders may include the bank's data science team, internal or
external auditors, regulators and, of course, the bank's customers. The
list of stakeholders may be larger for certain models.

\hypertarget{lifetime}{%
\subsubsection{Lifetime}\label{lifetime}}

In the next step it is necessary to define the life cycle of the model
and to identify which stakeholders are involved at which point. The role
of model developers is active at the beginning of the model life cycle.
The auditor's role begins when the model has been created. Then regular
monitorings and audits may be carried out periodically to determine the
current effectiveness of the model as required in \citep{fsb2017}.
Figure \ref{fig:xaiiceberg} presents a proposal, but for different
products the life cycle may vary.

\hypertarget{needs}{%
\subsubsection{Needs}\label{needs}}

After identifying the stakeholders and their activities during the model
life cycle, we can proceed to identify stakeholder needs. These needs
should meet the requirements listed in the regulations from the previous
section. Some examples

\begin{itemize}
\tightlist
\item
  A credit officer should be able to understand what were the key
  features behind a credit decision so that they can express his or her
  point of view and to contest the decision, as requested in
  \citep{EUdataregulations2018},
\item
  The auditor should be able to establish checks and balances at each
  stage of the development process as requested in \citep{fsb2017},
\item
  Proper oversight mechanisms shall be avaliable for auditor as
  requested in \citep{EthicsEU2019}.
\end{itemize}

\hypertarget{xai-methods}{%
\subsubsection{XAI methods}\label{xai-methods}}

There are different aspects in which one can categorise methods for
model interpretations. One is whatever method is model-specific,
i.e.~works only for a selected class of predictive models, or is
model-agnostic, that does not assume anything about the internal
structure of a model, such as e.g.~tracing signal through layers of deep
neural network \citep{montavon2018dsp}. Model specific tools can be very
powerful since they make use of the explicit model structure for the
analysis, e.g.~for linear models such as logistic regression the
interpretation is straightforward and given by the resulting effect
estimates. In this paper, we focus on model-agnostic tools. These tools
offer the opportunity to compare side by side models of different
structures, which allows for champion-challenger analysis of black box
models against interpretable glass box models.

The choice for a specific method should be based on the identified needs
of the stakeholder. In order to identify an appropriate set of
techniques the proposed pyramid of XAI methods depicts available methods
along two dimensions.

The first dimension - the vertical direction of the pyramid -
corresponds to the depth of model exploration. Methods are divided with
respect to the aspect of the model or prediction that should be
analysed. The underlying idea is to start with a simple measure of model
and prediction performance and drill down into factors that influence
model performance and individual predictions. It starts with general
performance metrics. Subsequently, they are broken down into components
related to specific characteristics of the credit application. The next
step is to analyse the response profile as ``what-if'' questions. This
cascade approach allows stakeholders to choose a level of explanation
that matches their needs.

The second (horizontal) dimension concerns the global versus local
aspect of model explainability, shown as the left and right side of the
pyramid. The global aspect is needed to verify the possible systemic
discrimination or biases of the model and to provide specific overall
checks. The local aspect is needed to facilitate explanation of the
model's for individual credit applications.

\begin{figure}

{\centering \includegraphics[width=1\linewidth]{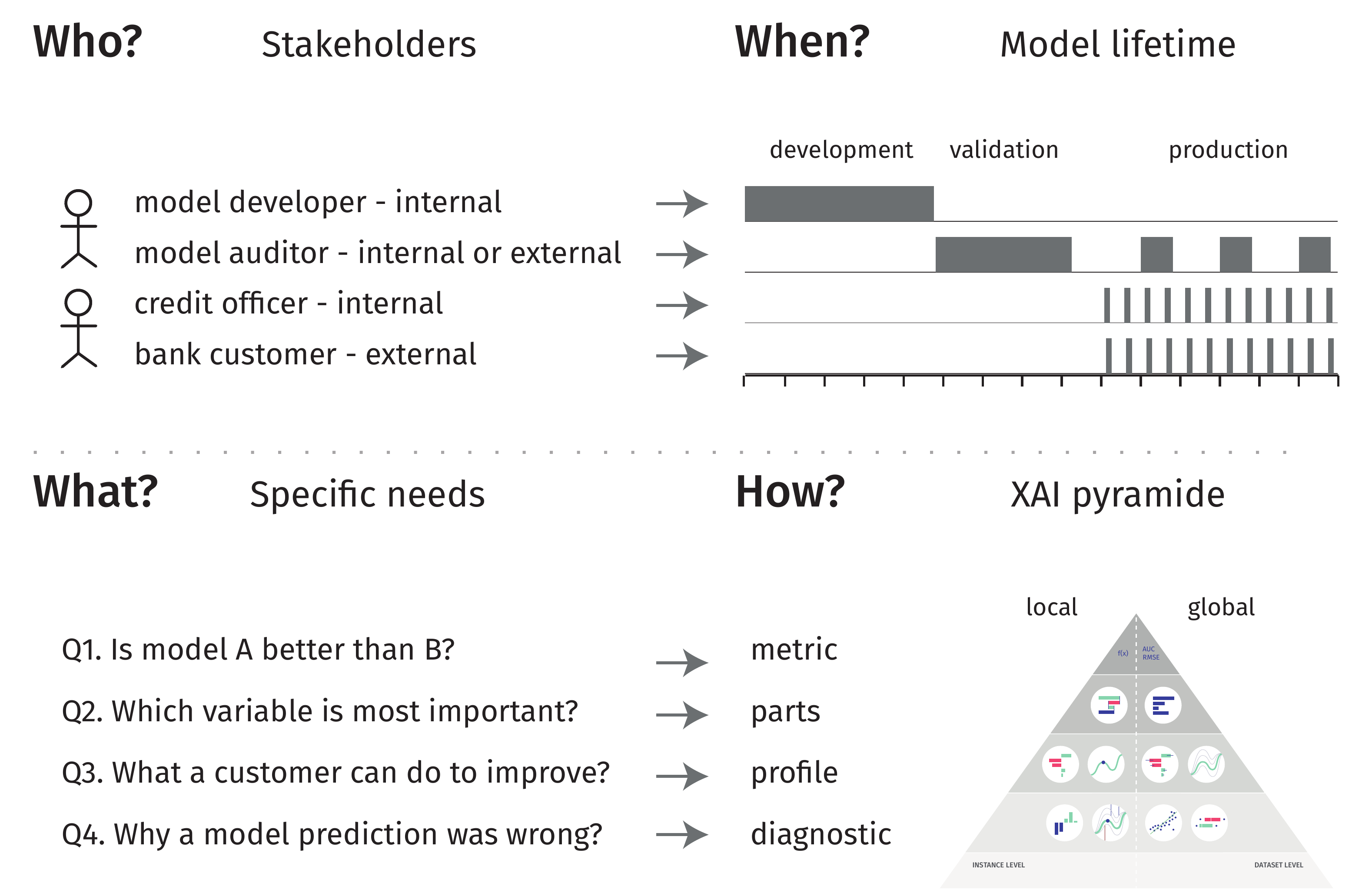} 

}

\caption{\label{fig:xaiiceberg} The process description consists of four components. The first defines the stakeholders that are affected by the model. The second defines the life cycle of the model and specifies which stakeholders are active in which part of the model life cycle. The third specifies at which time which stakeholders have what needs related to the model. The last component sets out the machine learning techniques that can be used to meet the identified needs.}\label{fig:xaiiceberg}
\end{figure}

We now summarize a list of methods for the two groups of methods for
local-level and global-level explanations
\citep{oreilly2018, mol2019int, pmvee2019, sze2019pre}. The methods
presented below are implemented in packages for R
\citep{bie2018dal, molnar2018iml} or Python \citep{interpret2019}.

\textbf{Global-level} explanations are focused on the model behavior in
general. One example is the model-agnostic feature importance. This can
be calculated as drop in the model performance after a permutation of a
selected feature \citep{figher2018vip}. Once the most important features
are identified, they can be examined in detail using Partial Dependence
Plots \citep[PDP,][]{Friedman00greedyfunction, greenwell207pdp}. PDPs
summarise how on average the model response changes with a shift of the
value for a single feature.

\textbf{Local-level} explanations are focused on a model behavior around
a single model prediction. Individual conditional expectations
\citep{goldstein2015peeking} trace how the model response would change
with a shift in the model input. The disadvantage of such an approach is
that one needs to trace every variable. This can be problematic for
large numbers of variables. Local Interpretable Model-agnostic
Explanation \citep[LIME,][]{lime} identifies sparse explanations based
on small numbers of features. LIME is based on simple interpretable
surrogate models that are fitted locally to the black box model. A
disadvantage of LIME is that feature explanations do not add to the
model predictions. In contrast, the SHapley Additive exPlanations (SHAP)
method \citep{NIPS2017_7062} uses Shapley values from cooperative game
theory to attribute additive feature effects to final model predictions.
The iBreakDown method \citep{2019arXiv190311420G} can be used for
identification of non-additive decompositions.

\hypertarget{comparative-study-of-scorecards-and-explainable-machine-learning}{%
\section{Comparative study of Scorecards and Explainable Machine
Learning}\label{comparative-study-of-scorecards-and-explainable-machine-learning}}

To illustrate the proposed process, in this section we show a
comparative analysis of several models for credit scoring. We compare
both the performance and, more importantly, the explainability of a
logistic regression model with modern machine learning models in a
credit scoring context. For the comparative study, a publicly available
data set has been used. This data set was provided by FICO, one of the
major credit bureaus in the United States \citep{fic2018xml}.

\hypertarget{description-of-the-data}{%
\subsection{Description of the data}\label{description-of-the-data}}

The data relate to the Home Equity Line of Credit (HELOC) with customers
requesting a credit line in the range of \$5,000 - \$150,000. The data
set has \(n = 10,459\) observations of 23 covariates and one binary
target variable. The task is to predict whether a consumer was ever more
than 90 days overdue within a 24 months period. The predictor variables
are all quantitative or categorical, and come from anonymised credit
bureau data.

Eleven additional dummy variables have been created after manual
inspection of the data in order to identify missing observations where
no information, no valid information or no bureau data is available. In
order to support rigorous analysis, the data set has been split and
randomly assigned into two sets: training data (75\% or 7844 obs.) and
test data (25\% or 2615 obs.), where the former has been used for model
training and parameter tuning while the latter has been set aside and
only used for performance evaluation.

In addition to creating a model of highest discriminatory power
according to the challenge description, a monotonic dependency of the
risk prediction on some of the predictor variables has to be ensured
\citep[cf.][]{fic2018xml}.

\hypertarget{models-for-comparison}{%
\subsection{Models for comparison}\label{models-for-comparison}}

\hypertarget{scorecard-model}{%
\subsubsection{Scorecard model}\label{scorecard-model}}

In the first step, a traditional credit risk scorecard has been
developed as a baseline for further comparison to represent the current
industry standard. A typical scorecard modelling process consists of a
chain of subsequent modelling steps and can be considered as a special
case of well-known general process standards for data mining business
practice such as KDD or CRISP-DM \citep{aze2008kdd}. Typically, at each
stage of the process, exploratory analyses are conducted resulting in
modelling decisions based on business knowledge \citep{sze2017afr}. A
typical preprocessing \citep[cf.~e.g.][]{fin2012cre} consists in
\emph{coarse classing} of the original variables as well as an optional
subsequent \emph{WOE transform} of the resulting dummy variables. The
preprocessd data are used for \emph{logistic regression} using
\emph{variable selection}, which is often performed by manual
interaction of the analyst using business information with suggestions
of an automatic selection strategy.

Variable coarse classing is generally obtained using an initial
algorithm-based binning which has to be manually updated by the analyst
in a second step. In the case of the HELOC data, the initial binning has
been generated using recursive binary splits of numeric variables that
maximise the information value
(\(IV = \sum_x (f(x|1)-f(x|0)) \, WOE(x)\)) of the binned variable,
where a minimum relative IV improvement of 5\% for any additional split
has to be reached. For categorical variables, all levels that do cover
less than 2\% of the data are assigned to a new \emph{``rest''} level.
The remaining levels are sorted according to their default rate and any
two subsequent levels are merged (from low to high) as long as their
default rates do not significantly differ (\(\alpha = 0.1\)) using a
\(\chi^2\) test \citep{sze2017afr}. Afterwards, few resulting very small
classes of numeric variables have been manually merged with their
adjacent class as well as a few bins that were violating the
monotonicity constraints as given by the competition
\citep[cf.][]{fic2018xml} after the initial automatic binning. An
example for this is given by the variable \emph{months since the most
recent delinquency}: As an advantage over the challenger black box
algorithms presented in the next section, the traditional approach of
white box scorecard development allows for a visualization of the
resulting bins in terms of default rates and by manual inspection in
this case a non-linear trend can be observed in the data (cf.~Figure
\ref{fig:binning}, left). A challenge now consists in explaining such
non-linearities to the stakeholders and regulators. In case of this
competition a constraint of monotonicity of the default rate w.r.t. the
variable \emph{months since the most recent delinquency} was required
and therefore the \(2^{nd}\) and \(3^{rd}\) class have been manually
merged (\ref{fig:binning}, right).

\begin{figure}

{\centering \includegraphics[width=1\linewidth]{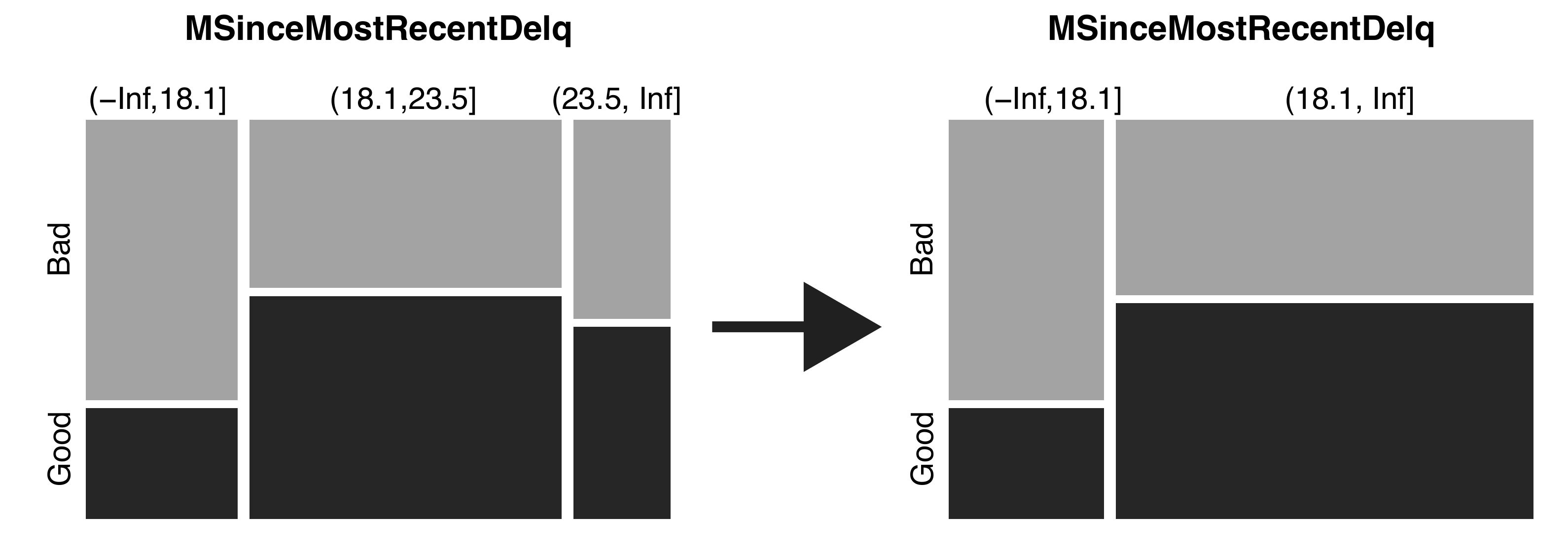} 

}

\caption{\label{fig:binning}Example: automatic vs. manual binning for the variable months since the most recent delinquency.}\label{fig:binning}
\end{figure}

The resulting coarse classes are given in the Appendix. Subsequently
WOEs are assigned to the classes:
\(WOE(x) = \log(\frac{f(x|1)}{f(x|0)})\) \citep[cf.~e.g.][]{tho2002cre}.
For reasons of parsimony and in order to ensure that the resulting model
respects the trends of the risk drivers in practice WOEs are typically
used instead of dummy variables which has also been the case for the
HELOC data.

The resulting WOE variables are used to train a logistic regression
model where, in a forward selection manner, variables are chosen using
marginal information values \citep[MIV,][]{sca2011clas} as long as the
improvement of the model as measured by the MIV is at least 0.01. The
resulting scorecard model has 16 variables and it is given in the
appendix. It has been scaled to a score of 500 for odds of 50 and 20
points to double the odds.

An advantage of the preceding methodology using coarse classing is that
it takes into account nonlinear relationships between the predictor
variables and the target, while simultaneously guaranteeing a
plausibility check by the analyst after each modelling step. This
plausibility check allows for the integration of business expert
experience, as required by the regulators. On the other hand, it is not
possible to cover nonlinear high-order multidimensional dependencies in
the data and this becomes a potential source of error resulting from the
manual interference if the number of variables is large. The latter two
issues can be overcome using modern machine learning techniques but the
resulting models typically are of a black box nature, as it has been
outlined in the previous section.

\hypertarget{challenger-approaches}{%
\subsubsection{Challenger Approaches}\label{challenger-approaches}}

Several black-box machine learning models have been tested to challenge
the scorecard model. The algorithms considered in this paper are
state-of-the-art machine learning models, including: generalised boosted
models \citep{gbm}, elastic net \citep{glmnet}, logistic regression with
spline-based transformations \citep{harrell2015regression}, two
implementations of random forests \citep{randomForest, ranger}, support
vector machines \citep{Cortes1995}, and extreme gradient boosting
\citep{Chen}. Moreover, two automated machine learning frameworks were
examined, namely H2O \citep{h2obook} and mljar \citep{mljar}. These
models are able to cover a wide range of potential relationships between
variables and can capture complex interactions.

The challenger models have been trained without any further data
preprocessing. Random search optimisation has been used to tune
hyperparameters of the models. All optimised hyperparameters and their
tuning ranges are provided in Table \ref{tab:challenger_hyperparameters}
and Table \ref{tab:challenger_hyperparameters_ranges}.

On the test data, the best results were obtained for the logistic
regression with spline-based transformations (rms), see Figure
\ref{fig:model_performance}. For 13 continuous variables the linear
tail-restricted cubic spline transformation was applied to adapt for
non-linear relations. Additionally, we also considered a model with
manually tuned penalty parameters. Reproducible code for all models can
be found in the respective GitHub repository:
\url{https://github.com/agosiewska/fico-experiments}.

\hypertarget{results-of-model-and-instance-level-exploration-and-explanation}{%
\section{Results of model and instance level exploration and
explanation}\label{results-of-model-and-instance-level-exploration-and-explanation}}

\hypertarget{model-level-exploration-and-explanation}{%
\subsection{Model level exploration and
explanation}\label{model-level-exploration-and-explanation}}

\hypertarget{model-performance}{%
\subsubsection{Model performance}\label{model-performance}}

For the model level exploration, the first step is usually related to
model performance. Different measures may be used. Common choices for
credit scoring include AUC, Gini, KS, F1, pAUC \citep{robin2011},
H-measure \citep{hand2009} or profit-based measures
\citep{crook2007, verbraken2014}. As the discussion of performance
measures is not the scope of this paper, we apply the commonly used AUC
for the remainder of the paper.

We denote the performance of a model \(f_\theta\) measured on a dataset
\(X\) with known values of a target variable \(y\) as

\[
M(\theta) = L(f_\theta, X, y),
\]

where \(L\) represents the performance measure (loss function) of
interest.

Typically, model performances on training and test data are compared. It
is common practice to compare model performance on at least
out-of-sample and out-of-time test data. In case of the HELOC data out
of time data is not available. For this reason, during the comparative
study of this paper, we restricted the analysis to a comparison of model
performance between training and test data.

A performance comparison of the various models is provided in Figure
\ref{fig:model_performance}. Surprisingly, the more complex machine
learning models showed only slightly superior performance compared to
the baseline scorecard using the HELOC data. The best results on both
training and test data were obtained for logistic regression using
spline transformations.

\begin{figure}

{\centering \includegraphics[width=0.7\linewidth]{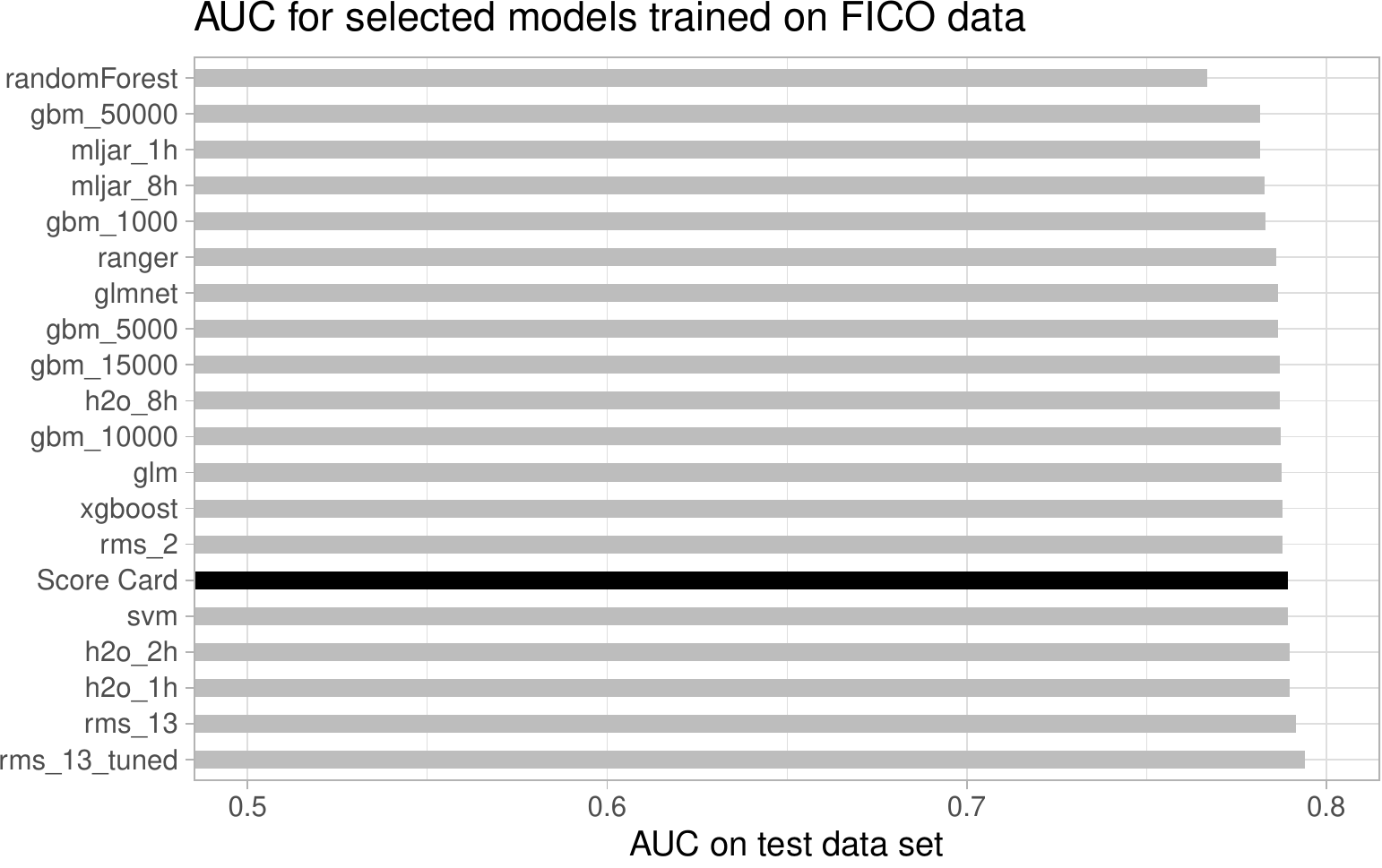} 

}

\caption{\label{fig:model_performance} Model performance measured by AUC on test data}\label{fig:model_performance}
\end{figure}

Elastic models are known to be sensitive to overfitting. Usually,
overfitting is assessed as the difference in model performance on the
training and test dataset.

\[
\text{overfitting}(\theta) = L(f_\theta, X_{\text{train}}, y_{\text{train}}) - L(f_\theta, X_{\text{test}}, y_{\text{test}}),
\]

Figure \ref{fig:model_overfitting} shows a scatterplot with model
performance on the train and test data sets. Note the different scale of
both axes. The graph nicely shows the common performance gap between
training and test data for random forests
\citep[cf.~e.g.][]{sze2017ont}. For further analysis we selected four
models based on a comparison of their predictive power on the training
and test data to account for potential overfitting (cf.~Figure
\ref{fig:model_overfitting}): the traditional scorecard, the SVM, the
GBM10000 as well as the best model.

\begin{figure}

{\centering \includegraphics[width=0.8\linewidth]{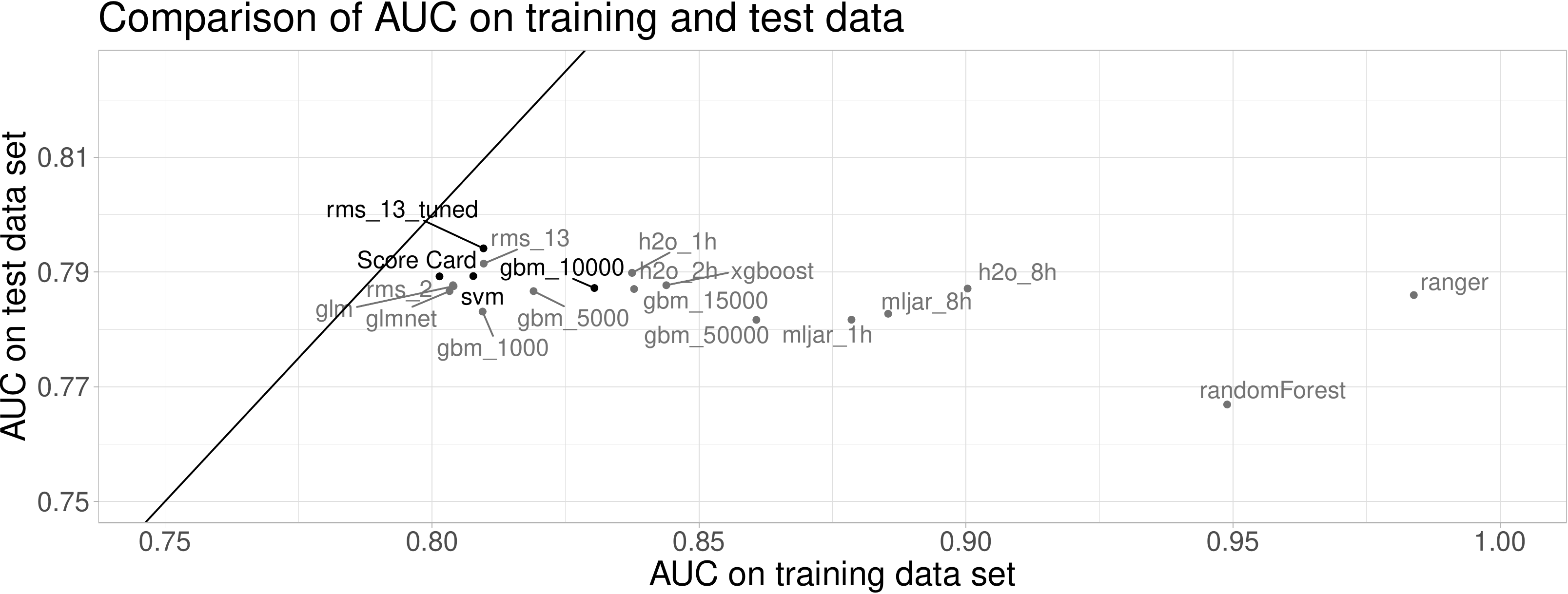} 

}

\caption{\label{fig:model_overfitting} Selection of models for comparison based on AUC on training and test data. Black line stands for AUC equal on training and test sets. The included models are gradient boosting machines (`gbm`) with different numbers of trees, logistic regression (`glm`), elastic net (`glmnet`), logistic regressions with spline based transformations (`rms`), two implementations of random forest (`randomForest`, `ranger`), support vector machines (`svm`), and extreme gradient boosting (`xgboost`). There are also included H2O's AutoML (`h2o`) and MLJAR AutoML (`mljar`) trained for different amount of time.}\label{fig:model_overfitting}
\end{figure}

\hypertarget{variable-importance}{%
\subsubsection{Variable importance}\label{variable-importance}}

The next step, according to the introduced framework, consists in
assessing the importance of each variable. Model agnostic variable
importance \citep{figher2018vip} measures how much the model's loss
function (or performance function) will change if a selected variable is
randomly permuted.

It may be simply introduced as

\[
FI(\theta, i, X, y) = L(f_\theta, X, y) - L(f_\theta, X^{*,j}, y),
\] where \(X^{*,j}\) is a dataset \(X\) with \(j^{th}\) column being
permuted.

Figure \ref{fig:ml_varimp} shows the importance \(FI\) for the GBM 10000
model in terms of a decrease in 1-AUC, which takes into account the
explicit performance measure under investigation. As an example, the
most important feature is the \texttt{ExternalRiskEstimate}: After
permutation of this variable the 1-AUC increases from 0.25 to over 0.28.

\begin{figure}
\includegraphics[width=0.8\linewidth]{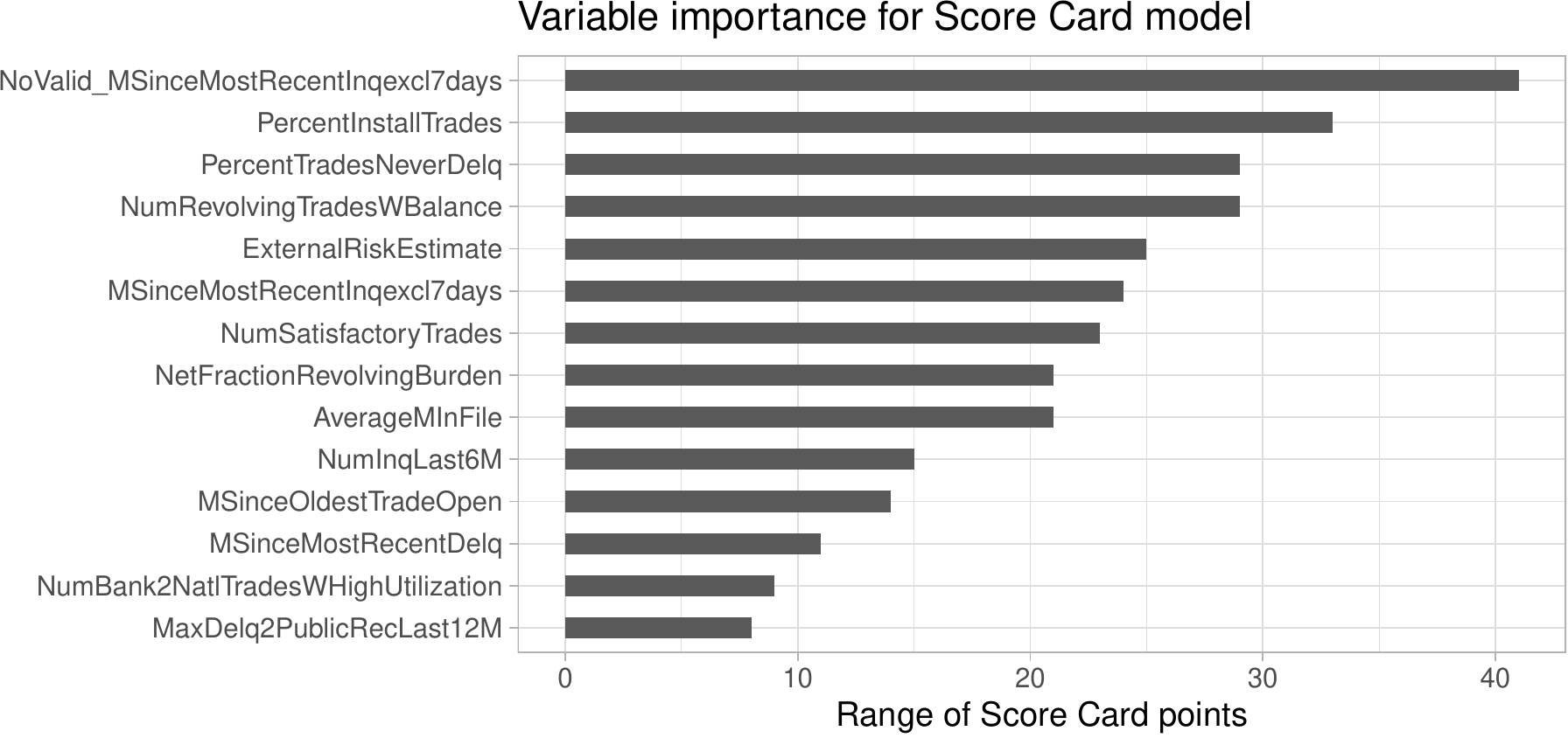} \caption{\label{fig:sc_varimp} Range of scorecard points as measure of variable importance for the Score Card}\label{fig:sc_varimp}
\end{figure}

\begin{figure}
\includegraphics[width=1\linewidth]{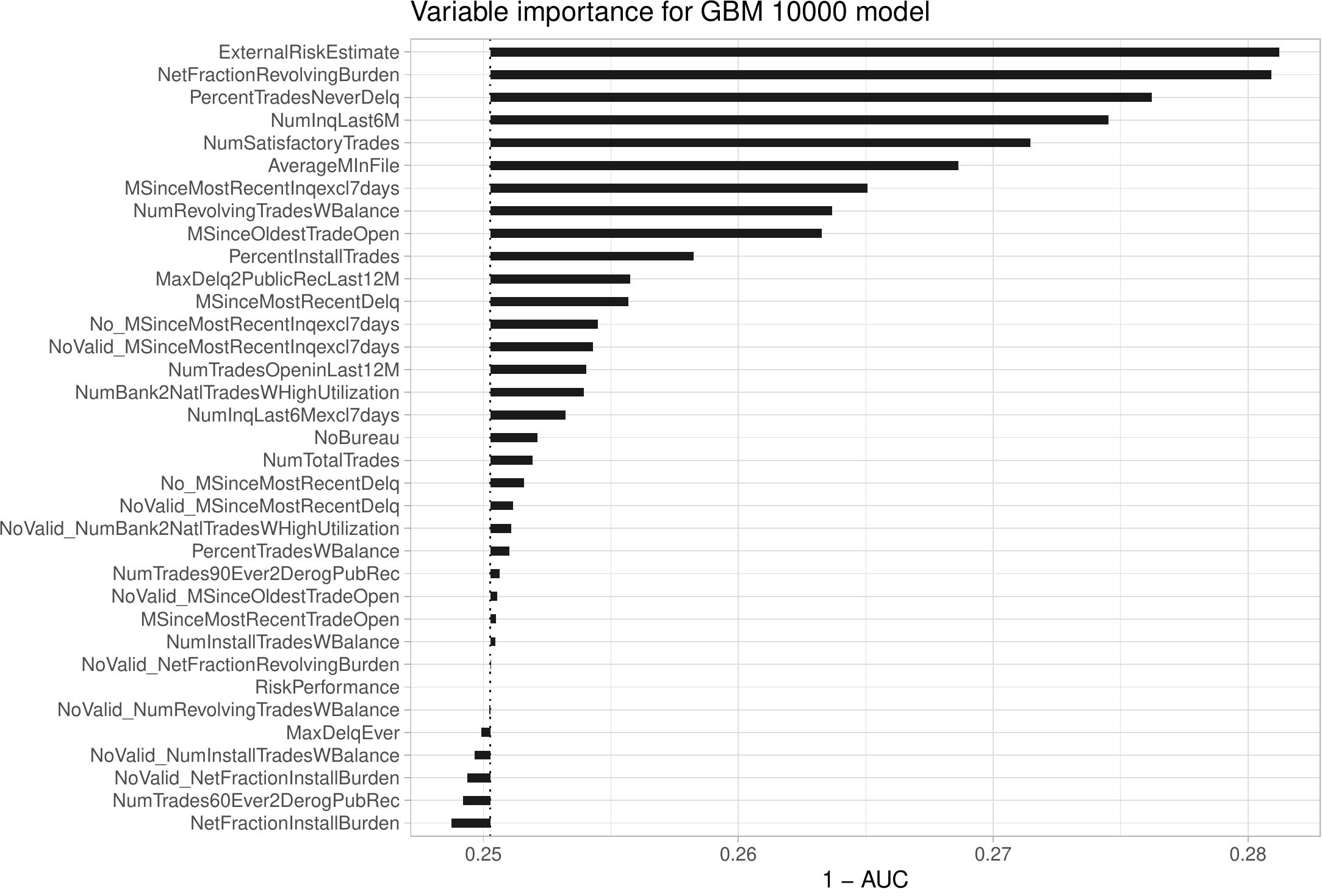} \caption{\label{fig:ml_varimp} Drop-out loss of AUC as measure of variable importance for the Machine Learning models.}\label{fig:ml_varimp}
\end{figure}

Note that variable importance corresponds to the effect of a variable if
all other variables are part of the model (similar to a backward
variable selection from a full model) which should be taken into account
during the analysis \citep{sze2019pre}.

\hypertarget{variable-effects}{%
\subsubsection{Variable effects}\label{variable-effects}}

Once the most important features are selected, the next layer of model
exploration is the assessment of marginal effects. Partial Dependency
Profiles are useful plots that show how an average model response
changes along changes in a selected feature.

PD profiles can be calculated as

\[
PD(\theta, i, z) = E[f_\theta(x|^{i}=z)]
\]

where \(x|^{i}=z\) is an observation \(x\) with \(i\)th coordinate
replaced by value \(z\). The \(E\) stands for expected value over the
marginal distribution of all variable except \(i\).

For the scorecard model (Figure \ref{fig:sc_marginal}) the effect of a
variable is directly given by the scorecard points (cf.~Appendix), which
are often a linear transformation of a variable's effect on the logit of
the default probabilities. For the Scorecard, the variables' effects are
given by step functions, which result from the preliminary coarse
classing step in model development.

\begin{figure}

{\centering \includegraphics[width=0.8\linewidth]{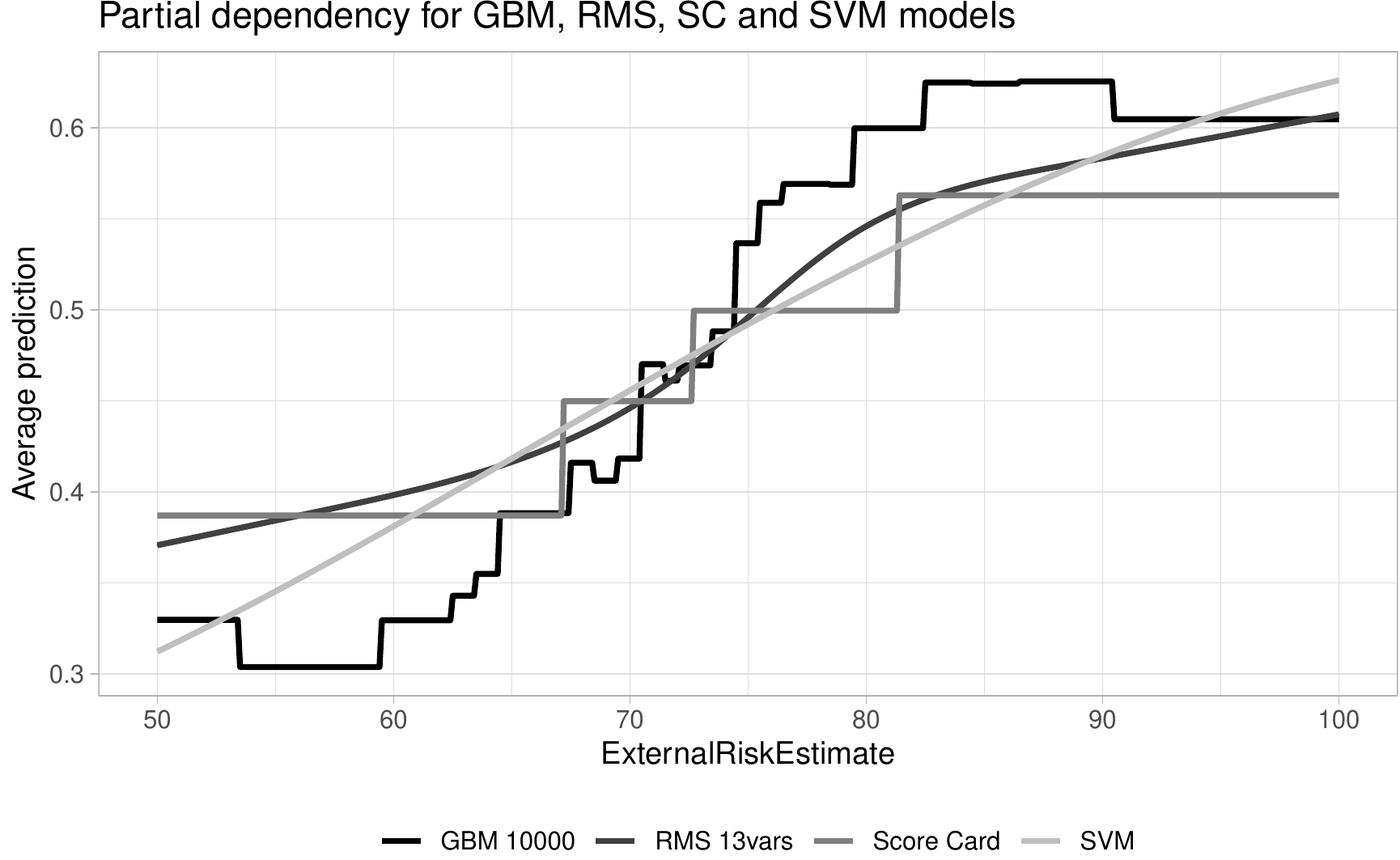} 

}

\caption{\label{fig:ml_marginal} Partial Dependence plots for the variable 'ExternalRiskEstimate' based on the selected Machine Learning models}\label{fig:unnamed-chunk-1}
\end{figure}

Figure \ref{fig:ml_marginal} compares all four models using PDPs for the
variable \texttt{ExternalRiskEstimate}. This suggests that more complex
models typically show smoother responses. Three of the presented models,
SVM, scorecard and RMS, have monotonic responses while the GBM model
exhibits some non-monotonic behavior on the edges.

\begin{figure}

{\centering \includegraphics[width=0.7\linewidth]{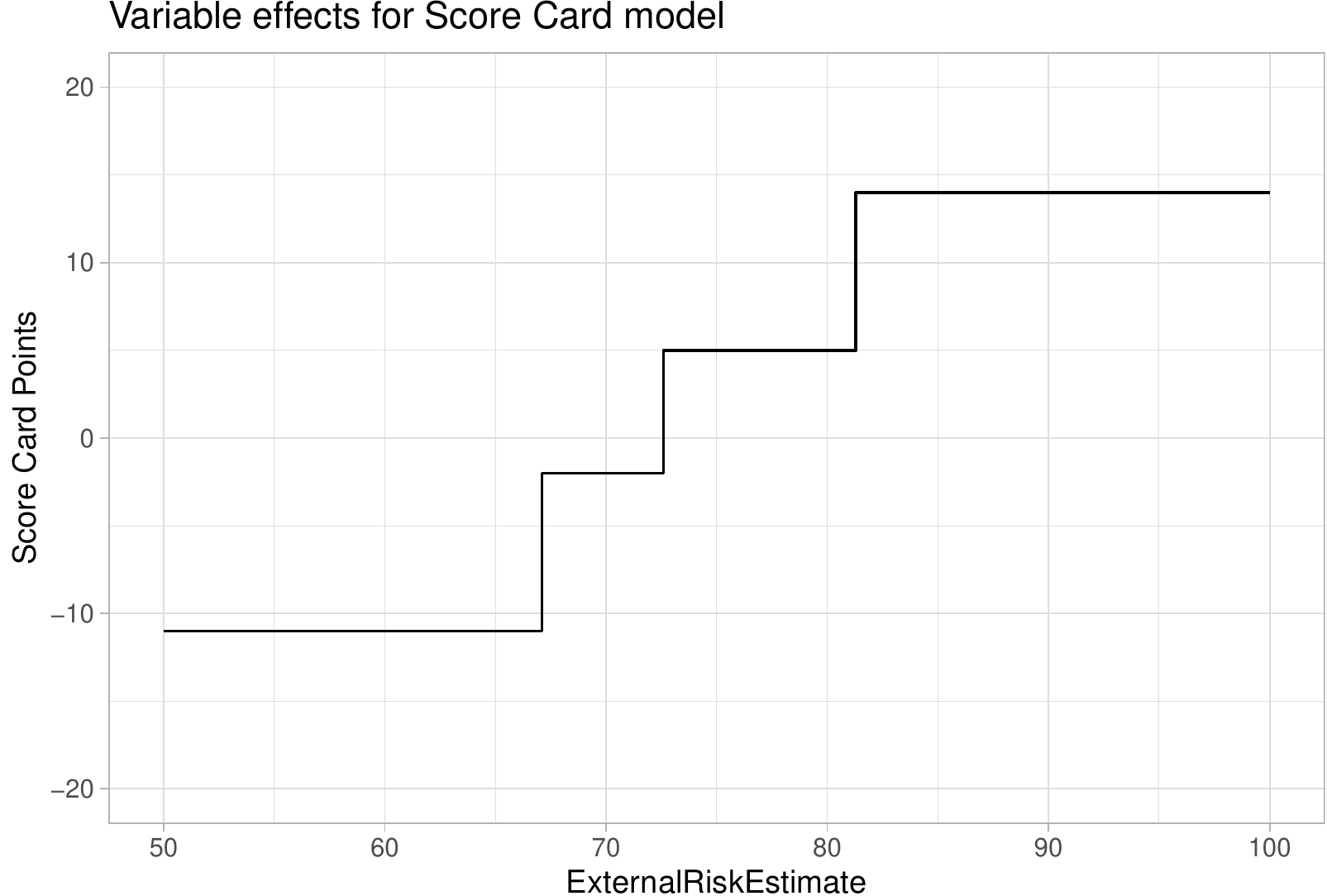} 

}

\caption{\label{fig:sc_marginal} Marginal effect of variable 'ExternalRiskEstimate' based on Score Card points}\label{fig:sc_marginal}
\end{figure}

Note that the Partial Dependency Profile is an average of individual
model responses. A sample of such individual responses is presented in
Figure \ref{fig:ml_cp}. If these individual profiles are parallel to
each other, then an average correctly describes individual behavior of
the model. In the presence of interactions, model responses may not be
parallel.

Whenever PDPs are used it has to be kept in mind that the value of a
true model prediction typically differs from the PDP depending on the
explicit values in all other variables. In \citep{sze2019how} a measure
is proposed to quantify how well the visualisation as given by a partial
dependence function matches the predictions of the model of interest.

\begin{figure}
\includegraphics[width=1\linewidth]{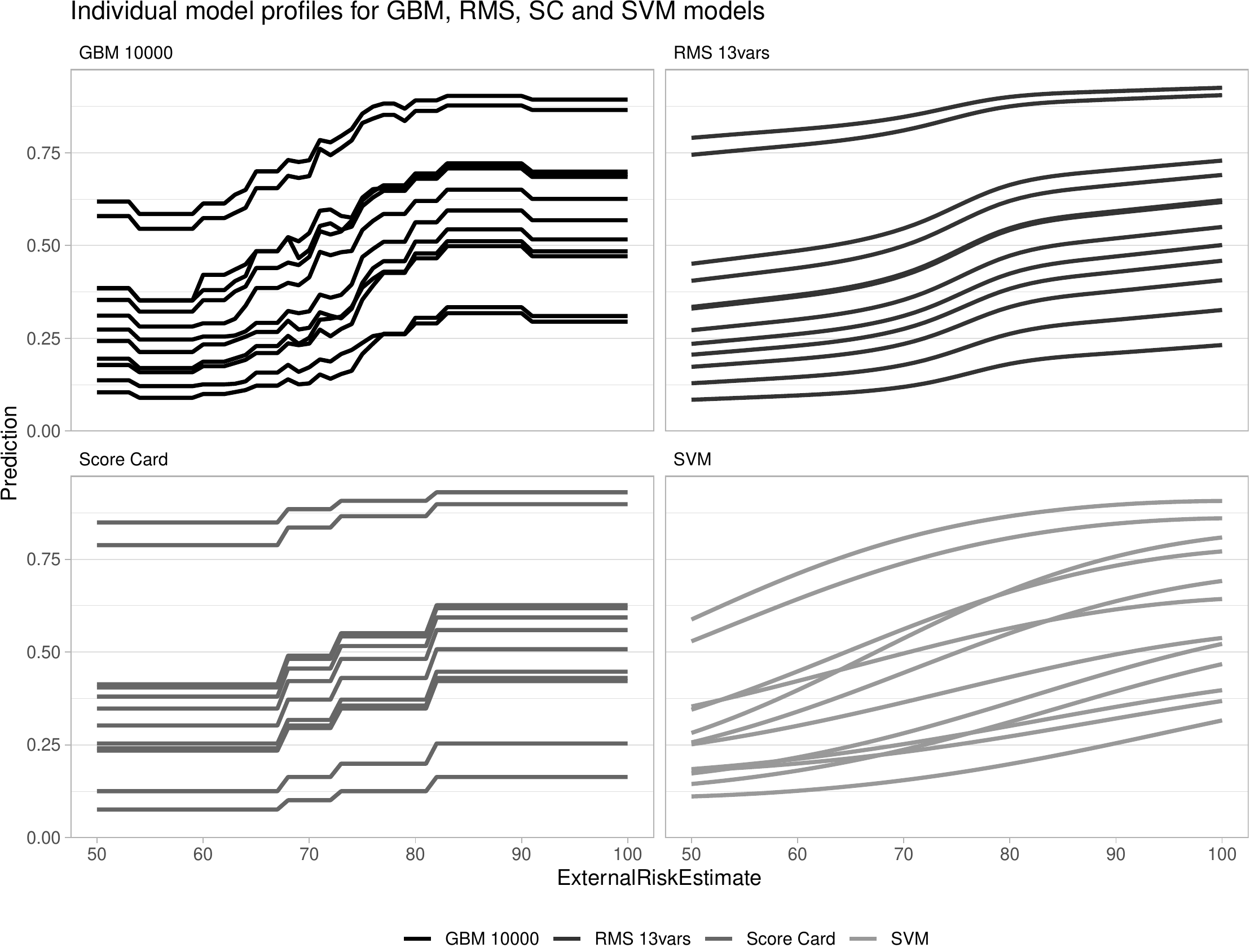} \caption{\label{fig:ml_cp} Ceteris Paribus plots for the variable 'ExternalRiskEstimate' based on the selected Machine Learning models}\label{fig:ml_cp}
\end{figure}

\hypertarget{instance-level-exploration-and-explanation}{%
\subsection{Instance level exploration and
explanation}\label{instance-level-exploration-and-explanation}}

For model level exploration, as presented in the previous section, the
average behavior of the whole model has been examined. For additive and
linear models, the average behavior is often similar to a local behavior
for particular instances. However, it is not necessarily the case for
more complex and elastic models. Interaction effects for particular
instances may be different than the average. For this reason in the
proposed framework, the subsequent step consists of instant level
explanations.

There are two primary use cases for instance level explanations. Either
the target is to explain the prediction for a new observation when the
model is applied (in this case we want to identify factors that
contribute strongly the model prediction) or a second possibility is
that some interesting points in the training data have been identified.
The latter case represents a useful tool for model debugging during the
development process: For the observations with largest residuals, one
may look for factors that fool the model predictions.

\hypertarget{model-prediction}{%
\subsubsection{Model prediction}\label{model-prediction}}

The first step of instance level explanations and exploration of a model
is to look at the accuracy of the prediction for a single observation.

The prediction is denoted as \(\hat y_i = f_\theta(x_i)\), where \(x_i\)
stands for \(i\)th observation.

\hypertarget{local-variable-attribution}{%
\subsubsection{Local variable
attribution}\label{local-variable-attribution}}

The second layer in the local part of the TAX4CS framework is given by
the attribution of particular features to the final model response for a
selected observation which allows to identify an applicant's most
adverse characteristics that were negatively contributing to a credit
rejection by a given model. For the traditional scorecard model, these
effects can be directly assessed by comparing scorecard points for a
single observation \(x\) in variable \(i\) to the average score of the
entire training sample in this variable. Figure \ref{fig:sc_instance}
shows an example where these differences are compared for all variables
and a single observation.

There are various model-agnostic methods which can be used to assign
local contributions to a single prediction, e.g.~LIME, SHAP or
iBreakDown as introduced in the previous section. A desired property for
such methods is the completeness of the explanation, i.e.~the sum of
variable attributions shall sum up to model response

\[
f_\theta(x) = \sum_i \delta_i(\theta, x),
\] where \(\delta_i(\theta, x)\) is the attribution of the \(i^{th}\)
variable. Both SHAP and iBreakDown do possess this property.

\begin{figure}
\includegraphics[width=1\linewidth]{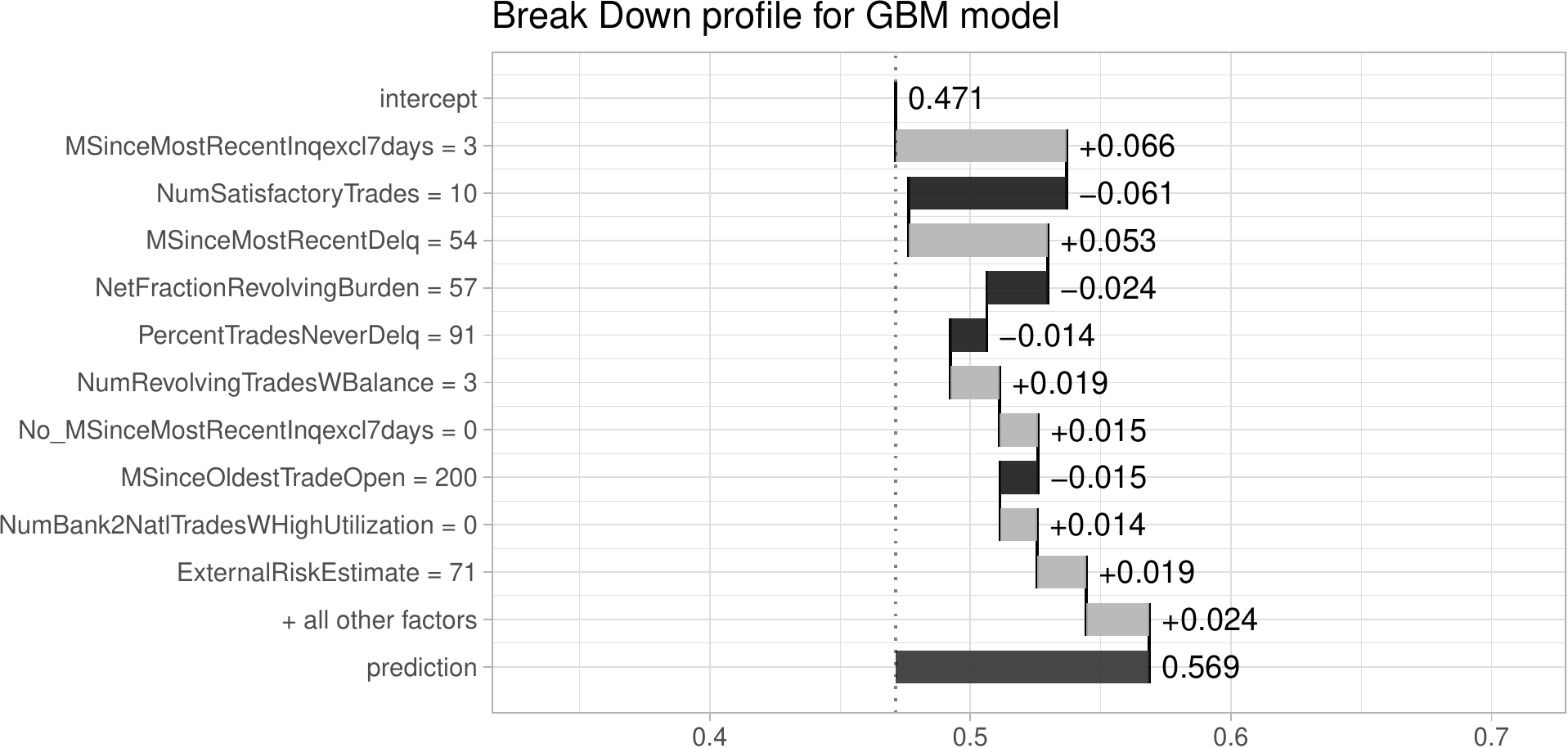} \caption{\label{fig:ml_addbreakdown} Additive breakdown for a single prediction as individual model agnostic explanation of variable importance for Machine Leaning models (here: Gradient Boosting)}\label{fig:unnamed-chunk-2}
\end{figure}

\begin{figure}
\includegraphics[width=1\linewidth]{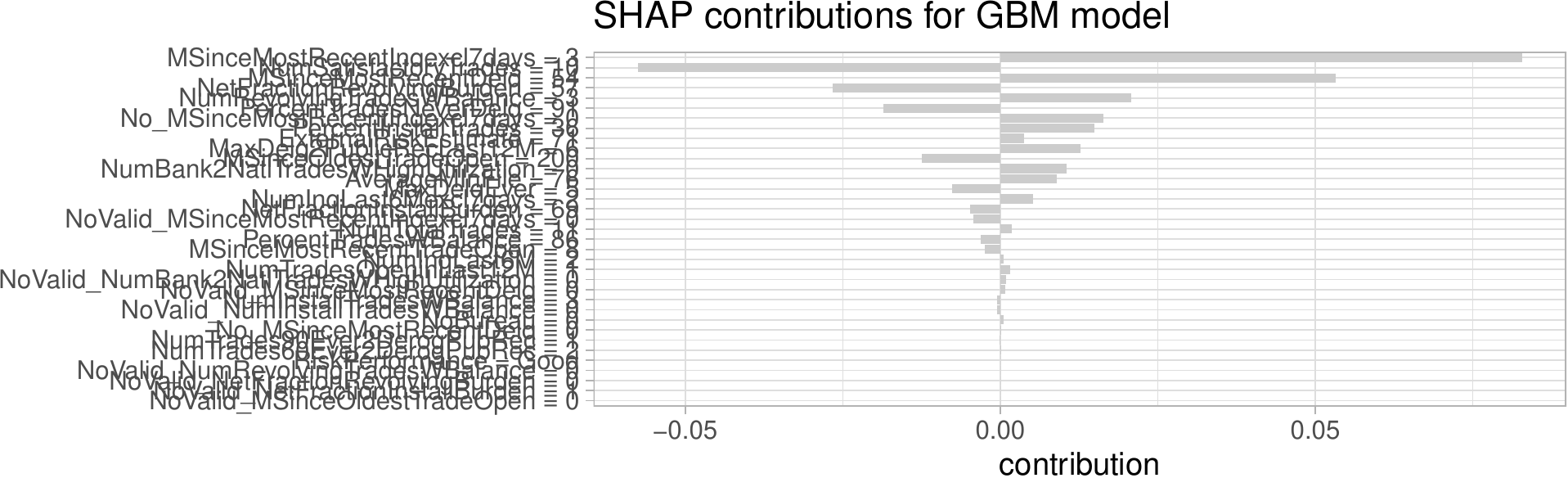} \caption{\label{fig:ml_shap} Average additive breakdown aka SHAP values for a single prediction as individual model agnostic explanation of variable importance for Machine Leaning models (here: Gradient Boosting)}\label{fig:unnamed-chunk-3}
\end{figure}

Figure \ref{fig:ml_addbreakdown} shows the iBreakDown result for
instance level attributions of the GBM 10000 model. The three most
influential variables are presented in the first three segments of the
waterfall plot. For non-additive models, feature attribution depends on
the order in which features are added to the iBreakDown Plot. Figure
\ref{fig:ml_shap} shows an average contribution from different paths.
Such averages are approximations of Shapley Additive exPlanations (SHAP
values).

The main goal of these attributions is to identify key factors that
influence model prediction.

\hypertarget{local-variable-effects}{%
\subsubsection{Local variable effects}\label{local-variable-effects}}

The third layer of the local part of TAX4CS is also related to effects
of particular features on model response for a single observation. These
effects can be captured with Ceteris Paribus (CP) profiles. Figure
\ref{fig:ml_cp} shows responses for the \texttt{ExternalRiskEstimate}
for 10 example observations.

CP profiles for an observation \(x\) and a variable \(i\) are defined in
the following way:

\[
CP(\theta, x, i, z) = f_\theta(x|^{i}=z),
\]

i.e.~it corresponds to a model response profile for an observation in
which variable \(i^{th}\) is changed to \(z\).

Please note that the partial dependency profile is a pointwise average
of individual Ceteris Paribus profiles. For additive relations,
individual Ceteris Paribus profiles are parallel and have the same shape
as the Partial Dependency profile. However, for non-additive models,
individual profiles can bring additional information about
instance-specific model behavior.

\begin{figure}
\includegraphics[width=1\linewidth]{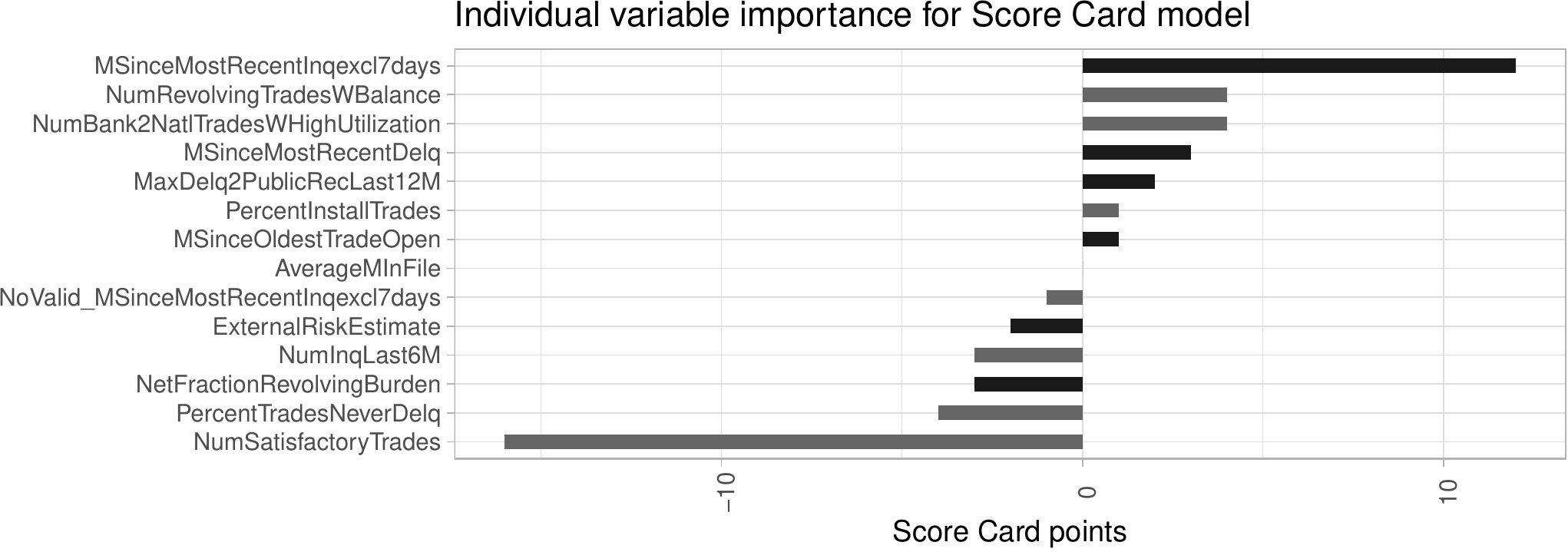} \caption{\label{fig:sc_instance} ScorecardPoints for a single prediction as individual explanation of variable importance}\label{fig:sc_instance}
\end{figure}

\hypertarget{conclusions}{%
\subsection{Conclusions}\label{conclusions}}

In this section, we saw a minor advantage of complex machine learning
models over the the traditional scorecard baseline model. Similar
results were also described by participants in the Explainable Machine
Learning Challenge. For example, the second-place winners have shown
that an SVM model with a linear kernel achieved higher accuracy than
more complex models, such as random forest \citep{holteretal}. Also when
comparing multiple measures, such as balanced and unbalanced accuracy,
AUC, and Kolmogorov Smirnov statistic complex models have achieved
little or no advantage over simple models \citep{9050779}. As a first
conclusion we want to note that a blind belief in the superior
performance of modern ML algorithms should always be challenged and a
proper benchmark analysis of its benefits from the stakeholder's
perspective should be undertaken for each separate scorecard model
development \citep[see also][]{sze2017ont, Rudin2019}.

Many participants have developed new explainable classification
algorithms: The winning team proposed the column generation (CG)
algorithm to efficiently search the best possible rule set
\citep{10.5555/3327345.3327376}, while the Recognition Award winning
team used 10 small regression models combined into a globally
interpretable model \citep{chenetal}.

The teams that took high spots in the competition used existing
interpretable models or developed their own one. The suggested
\textbf{framework for Transparency, Auditability and eXplainability for
Credit Scoring} (TAX4CS) provides a structured set of steps required for
explanatory analysis of any complex model, which makes it possible to
place greater emphasis on the \emph{suitability} of the model instead of
\emph{interpretability} only. It allows to not only explore a single
model but also provides tools for comparison of several models. Once the
algorithms of interest are chosen, the similarity of them might be
assessed by variable importance. Models driven by similar factors would
have an overlap between the most important variables. In the next step,
comparing PD profiles of important variables between models can be used
to determine whether the type of relationships between variables and
predictions are the same or different for different models, for example,
linear vs squared relationships. In the HELOC dataset example, SVM,
scorecard, and RMS models had similar monotonic responses while the GBM
model was non-monotonic at the edges. With this insight, it might be
desired to further check of suitability of GBM, whether it caught a
relationship undiscovered by other models or overfitted due to the small
number of observations with extreme values. In addition, PD profiles can
be used to assess the model's appropriateness. If the dependency between
variable and model predictions is inconsistent with the domain knowledge
it can be considered as a signal that the model might be poorly suited.

An important aspect to be considered within scorecard developments is
\emph{bias}. As an example consider a binary variable denoting whether
one or several debtors are liable. From historical data it might turn
out that additional debtors were added often if the creditworthiness of
the applicant alone was not sufficient for granting the credit
(i.e.~more likely to occur in situations of a higher probability of
default). This kind of sampling bias will erroneously lead to an
increased score if a potential additional debtor will be removed from an
application which is of course not desirable from the business point of
view. In contrast, it is desirable to avoid such kind of wrong
conclusions from correlations in data which are the topic of causal
inference \citep[cf.~also][]{lueb2020}. A similar sampling bias can be
implied through reject inference \citep{banasik2007, buc2013rej}.
\citet{schoelkopf2019} reviews the impact on causal inference on machine
learning. The connection between partial dependence plots and the
back-door adjustment for causal effects is shown in \citet{zhao2019}:
PDPs can be used for causal inference if corresponding assumptions are
met. \citet{Gomez2020} suggests to use \emph{Visual Counterfactual
Explanations} (ViCE).

The analysis of a single observation can be performed with instance
level methods. The local variable attributions assess the influence of
the variables on the prediction for the selected observation. The
iBreakDown method was used for explaining GBM model in the HELOC data
set example because in contrast to LIME or SHAP, iBreakDown might be
supported by the analysis of the stability, presented in Figure
\ref{fig:ml_shap}. Once the contribution of variables is established,
the Ceteris Paribus (CP) profiles for the most impactful variables can
be used for what-if analysis of the model's predictions. The results
might be then confronted with domain experts to assess the suitability
of predictions.

The methods presented in this chapter with the example on the HELOC
dataset form a structured framework for explanation analysis of machine
learning models. The proposed framework provides tools to not only
assess the suitability of models but also to compare them. This is a
strongly increasing field of research in AI. The framework shows how
this can be added to the process for developing score cards. Future
research could look into how to include more and newly developed methods
into the framework.

\hypertarget{summary}{%
\section{Summary}\label{summary}}

We have demonstrated that interpretations comparable to those of
traditional logistic regression models are also possible for modern
complex machine learning techniques. We propose a structured framework
(TAX4CS) for model-level and instance-level exploration, starting with
general measures of model performance (or accuracy of single predictions
respectively) and drill down into detailed descriptions of model
behavior through variable importance and effects (attribution and
response profiles for instance level explanations respectively). For
every single step in the process we provide an introduction to model
agnostic measures and approaches that can be used for arbitrary
predictive models such as black-box Machine Learning algorithms used for
classification or regression. This framework can be used in fields of
application such as credit scoring as a guideline to ensure that the
required degree of explanability can be achieved.

In an empirical study on a publicly available credit bureau data set,
the available methodology for model transparency is compared to the
interpretability given by the traditional scorecard modelling approach.
Notably, the basic scorecard shows suprisingly good performance in
comparison with advanced Machine Learning techniques such as Gradient
Boosting or Support Vector Machines. As a consequence, we find that in
practice, it is advisable to run different models of different
complexity and carefully evaluate up to which degree a higher model
complexity is beneficial for each specific situation
\citep[cf.~also][]{sze2017ont}

The comparable performance of the scorecard model in our study can be
explained by the simple tabular structure of the data set as well as the
thorough manual data preparation that enables the Logistic Regression
model to capture the relevant information in a similar fashion as more
complex and non-linear models. Given that the use of transaction data
and additional external data sources (e.g.~in social scoring) is going
to increase significantly in the future, the data used for credit
scoring will become more complex and feature engineering will become
more important. Since manual data preparation for using logistic
regression with a large number of variables will become more and more
extensive and costly, machine learning will be able to leverage its
strengths \citep{tobback2019, fsb2017}. In this case, the presented
model exploration process will be inevitable in order to meet regulatory
requirements.

\hypertarget{appendix-scorecard-model}{%
\section{Appendix: Scorecard Model}\label{appendix-scorecard-model}}

\begin{table}[H] \tiny
\caption{Scorecard model.}

{\begin{tabular}{lrrrr}
\toprule
Level               & Points      & \% Population & \% Default & Average PD \\
\midrule
Total population    &             & 1.00          & 0.52       & 0.521      \\             
\midrule
Intercept           & 385         &               &            & \\
\midrule
\multicolumn{5}{l}{ExternalRiskEstimate} \\
\midrule
(-Inf,67.1]         & -11         & 0.333         & 0.78       & 0.78 \\
(67.1,72.6]         &  -2         & 0.216         & 0.58       & 0.584 \\
(72.6,81.3]         &   5         & 0.253         & 0.39       & 0.388 \\
(81.3, Inf]         &  14         & 0.196         & 0.19       & 0.186 \\
\midrule
\multicolumn{5}{l}{AverageMInFile} \\
\midrule
(-Inf,59.3]         & -13         & 0.259         & 0.71       & 0.709 \\
(59.3,80.4]         &   0         & 0.327         & 0.52       & 0.527 \\
(80.4, Inf]         &   8         & 0.412         & 0.40       & 0.399 \\
\midrule
\multicolumn{5}{l}{NetFractionRevolvingBurden} \\
\midrule
(59.4, Inf]         & -13         & 0.204         & 0.78       & 0.762 \\
(26.3,59.4]         &  -3         & 0.359         & 0.59       & 0.602 \\
(-Inf,26.3]         &   8         & 0.436         & 0.35       & 0.342 \\
\midrule
\multicolumn{5}{l}{PercentTradesNeverDelq} \\
\midrule
(-Inf,58.4]         & -24         & 0.204         & 0.90       & 0.877 \\
(58.4,83.2]         & -12         & 0.131         & 0.75       & 0.754 \\
(83.2,95.4]         &  -4         & 0.308         & 0.60       & 0.602 \\
(95.4, Inf]         &   5         & 0.534         & 0.40       & 0.401 \\
\midrule
\multicolumn{5}{l}{MSinceMostRecentInqexcl7days} \\
\midrule
(-Inf,2.68]         &  -3         & 0.813         & 0.56       & 0.560 \\
(2.68,10.5]         &  12         & 0.129         & 0.38       & 0.387 \\
(10.5, Inf]         &  21         & 0.057         & 0.29       & 0.284 \\
\midrule
\multicolumn{5}{l}{NoValidMSinceMostRecentInqexcl7days} \\
\midrule
0         &  -1         & 0.951         & 0.54       & 0.539 \\
1         &  30         & 0.048         & 0.18       & 0.183 \\
\midrule
\multicolumn{5}{l}{MSinceMostRecentDelq} \\
\midrule
(-Inf,18.1]         &  -8         & 0.270         & 0.72       & 0.717 \\
(18.1, Inf]         &   3         & 0.729         & 0.45       & 0.449 \\
\midrule
\multicolumn{5}{l}{NumSatisfactoryTrades} \\
\midrule
(-Inf,11.2]         & -16         & 0.188         & 0.67       & 0.674 \\
(11.2,23.1]         &   0         & 0.460         & 0.52       & 0.509 \\
(23.1, Inf]         &   7         & 0.351         & 0.45       & 0.456 \\
\midrule
\multicolumn{5}{l}{NumBank2NatlTradesWHighUtilization} \\
\midrule
(2.4, Inf]          &  -5         & 0.120         & 0.72       & 0.713 \\
(0.383,2.4]         &  -2         & 0.469         & 0.61       & 0.609 \\
(-Inf,0.383]        &   4         & 0.419         & 0.37       & 0.365 \\
\midrule
\multicolumn{5}{l}{MSinceOldestTradeOpen} \\
\midrule
(-Inf,87.2]         &  -9         & 0.099         & 0.76       & 0.756 \\
(87.2,134]          &  -4         & 0.128         & 0.64       & 0.646 \\
(134,266]           &   1         & 0.566         & 0.51       & 0.506 \\
(266, Inf]          &   5         & 0.206         & 0.38       & 0.375 \\
\midrule
\multicolumn{5}{l}{PercentInstallTrades} \\
\midrule
(85.4, Inf]         &  -29         & 0.011         & 0.87       & 0.888 \\
(50.3,85.4]         &   -8         & 0.142         & 0.64       & 0.635 \\
(45.1,50.3]         &   -5         & 0.075         & 0.59       & 0.589 \\
(23.1,45.1]         &    1         & 0.492         & 0.50       & 0.510 \\
(-Inf,23.1]         &    4         & 0.278         & 0.46       & 0.451 \\
\midrule
\multicolumn{5}{l}{NumRevolvingTradesWBalance} \\
\midrule
(13.3, Inf]         &   -9         & 0.012         & 0.65       & 0.657 \\
(11.8,13.3]         &  -25         & 0.011         & 0.82       & 0.775 \\
(5.35,11.8]         &   -7         & 0.206         & 0.61       & 0.605 \\
(3.07,5.35]         &   -1         & 0.294         & 0.54       & 0.561 \\
(-Inf,3.07]         &    4         & 0.475         & 0.46       & 0.451 \\
\midrule
\multicolumn{5}{l}{NumInqLast6M} \\
\midrule
(6.13, Inf]        & -15         & 0.025         & 0.78       & 0.742 \\
(1.81,6.13]        &  -3         & 0.299         & 0.59       & 0.602 \\
(-Inf,1.81]        &   2         & 0.674         & 0.48       & 0.477 \\
\midrule
\multicolumn{5}{l}{MaxDelq2PublicRecLast12M} \\
\midrule
(-Inf,5.26]        &  -6         & 0.237         & 0.73       & 0.730 \\
(5.26, Inf]        &   2         & 0.762         & 0.46       & 0.456 \\
\bottomrule
\end{tabular}}

\label{sample-table}
\end{table}

\hypertarget{appendix-challenger-models}{%
\section{Appendix: Challenger Models}\label{appendix-challenger-models}}

The reproducible R and Python code for model tuning can be found in the
GitHub repository: \url{https://github.com/agosiewska/fico-experiments}.

\begin{table}[H] \tiny
\caption{Ranges of hyperparameters for challenger models.}
{\begin{tabular}{lll}
\textbf{Algorithm} & \textbf{Hyperparameter} & \textbf{Values}   \\ \hline
Generalised Boosted Models (gbm)                            & \multicolumn{2}{l}{}                        \\
                                                            & n.trees                 & {[}1000, 50000{]} \\ \hline
Logistic Regression (glm)                                   & \multicolumn{2}{l}{}                        \\ \hline
Elastic Net (glmnet)                                        &                         &                   \\
                                                            & alpha                   & {[}0,1{]}         \\
                                                            & lambda                  & {[}0, Inf{]}      \\ \hline
H2O AutoML                                                  & \multicolumn{2}{l}{}                        \\
                                                            & time                    & 1h, 2h, 8h        \\ \hline
MLJAR AutoML                                                &                         &                   \\
                                                            & time                    & 1h, 2h            \\ \hline
Random Forest (randomForest)                                &                         &                   \\
                                                            & ntree                   & {[}100, 5000{]}   \\
                                                            & mtry                    & {[}3, 20{]}       \\ \hline
Random Forest (ranger)                                      & \multicolumn{2}{l}{}                        \\
                                                            & num.trees               & {[}100, 5000{]}   \\
                                                            & mtry                    & {[}3, 20{]}       \\ \hline
Logistic Regression with Splines (rms) & \multicolumn{2}{l}{}                        \\
                                                            & manually tuned          &                   \\ \hline
Radial Basis Function kernel Support Vector Machines (svm)                               & \multicolumn{2}{l}{}                        \\
                                                            & cost                    & {[}0, Inf{]}      \\
                                                            & gamma                   & {[}0, Inf{]}      \\ \hline
Extreme Gradient Boosting (xgboost)                         &                         &                   \\
                                                            & eta                     & {[}0,1{]}         \\
                                                            & max\_depth              & {[}3, 10{]}       \\
                                                            & nrounds                 & {[}50, 2000{]}    \\
                                                            & lambda                  & {[}0, Inf{]}      \\
                                                            & alpha                   & {[}0, Inf{]}  \\ \hline  
\end{tabular}}
\label{tab:challenger_hyperparameters_ranges}
\end{table}

\begin{table}[H] \tiny
\caption{Optimal hyperparameters fo challenger models.}
{\begin{tabular}{llll}
Name         & \textbf{Algorithm}                  & \textbf{Hyperparameter} & \textbf{Value} \\ \hline
gbm\_100     & Generalised Boosted Models (gbm)    & \multicolumn{2}{l}{}                     \\
             &                                     & n.trees                 & 1000           \\
             &                                     & interaction.depth       & 3              \\ \hline
gbm\_5000    & Generalised Boosted Models (gbm)    &                         &                \\
             &                                     & n.trees                 & 5000           \\
             &                                     & interaction.depth       & 3              \\ \hline
gbm\_10000   & Generalised Boosted Models (gbm)    &                         &                \\
             &                                     & n.trees                 & 10000          \\
             &                                     & interaction.depth       & 3              \\ \hline
gbm\_15000   & Generalised Boosted Models (gbm)    &                         &                \\
             &                                     & n.trees                 & 15000          \\
             &                                     & interaction.depth       & 3              \\ \hline
gbm\_5000    & Generalised Boosted Models (gbm)    &                         &                \\
             &                                     & n.trees                 & 50000          \\
             &                                     & interaction.depth       & 3              \\ \hline
glmnet       & Elastic Net (glmnet)                &                         &                \\
             &                                     & alpha                   & 0.3881912      \\
             &                                     & lambda                  & 0.001120922    \\ \hline
randomForest & Random Forest (randomForest)        &                         &                \\
             &                                     & ntree                   & 296            \\
             &                                     & mtry                    & 3              \\ \hline
ranger       & Random Forest (ranger)              & \multicolumn{2}{l}{}                     \\
             &                                     & num.trees               & 1425           \\
             &                                     & mtry                    & 3              \\ \hline
svm          & Radial Basis Function kernel Support Vector Machines (svm)       & \multicolumn{2}{l}{}                     \\
             &                                     & cost                    & 0.2010253      \\
             &                                     & gamma                   & 0.01930817     \\ \hline
xgboost      & Extreme Gradient Boosting (xgboost) &                         &                \\
             &                                     & eta                     & 0.1583868      \\
             &                                     & max\_depth              & 5              \\
             &                                     & nrounds                 & 1515           \\
             &                                     & lambda                  & 185.8683       \\
             &                                     & alpha                   & 14.11926\\ \hline      
\end{tabular}}
\label{tab:challenger_hyperparameters}
\end{table}

\clearpage

\bibliographystyle{tfcad}
\bibliography{bib.bib}

\begin{thebibliography}{}

\bibitem[\protect\astroncite{Alemzadeh
  et~al.}{2016}]{10.1371/journal.pone.0151470}
Alemzadeh, H., Raman, J., Leveson, N., Kalbarczyk, Z., and Iyer, R.~K. (2016).
\newblock Adverse events in robotic surgery: A retrospective study of 14 years
  of fda data.
\newblock {\em PLOS ONE}, 11(4):1--20.

\bibitem[\protect\astroncite{{Ariza-Garzón} et~al.}{2020}]{9050779}
{Ariza-Garzón}, M.~J., {Arroyo}, J., {Caparrini}, A., and {Segovia-Vargas}, M.
  (2020).
\newblock Explainability of a machine learning granting scoring model in
  peer-to-peer lending.
\newblock {\em IEEE Access}, 8:64873--64890.

\bibitem[\protect\astroncite{Arya et~al.}{2019}]{arya2019explanation}
Arya, V., Bellamy, R. K.~E., Chen, P.-Y., Dhurandhar, A., Hind, M., Hoffman,
  S.~C., Houde, S., Liao, Q.~V., Luss, R., Mojsilović, A., Mourad, S.,
  Pedemonte, P., Raghavendra, R., Richards, J., Sattigeri, P., Shanmugam, K.,
  Singh, M., Varshney, K.~R., Wei, D., and Zhang, Y. (2019).
\newblock One explanation does not fit all: A toolkit and taxonomy of ai
  explainability techniques.

\bibitem[\protect\astroncite{Azevedo and Santos}{2008}]{aze2008kdd}
Azevedo, A. and Santos, M.~F. (2008).
\newblock Kdd, semma and crisp-dm: a parallel overview.
\newblock In {\em Europ. Conf. Data Mining (IADIS)}, page 182–185.

\bibitem[\protect\astroncite{Baesens et~al.}{2002}]{baes02ben}
Baesens, B., Gestel, T.~V., Viaene, S., Stepanova, M., Suykens, J., and
  Vanthienen, J. (2002).
\newblock Benchmarking state-of-the-art classification algorithms for credit
  scoring.
\newblock {\em JORS}, 54(6):627--635.

\bibitem[\protect\astroncite{Banasik and Crook}{2007}]{banasik2007}
Banasik, J. and Crook, J. (2007).
\newblock Reject inference, augmentation and sample selection.
\newblock {\em European Journal of Operational Research}, 183:1582--1594.

\bibitem[\protect\astroncite{Bellotti and Crook}{2009}]{belotti2009}
Bellotti, T. and Crook, J. (2009).
\newblock Support vector machines for credit scoring and discovery of
  significant features.
\newblock {\em Expert Systems with Applications}, 2(33):3302--3308.

\bibitem[\protect\astroncite{Biecek}{2018}]{bie2018dal}
Biecek, P. (2018).
\newblock Dalex: explainers for complex predictive models.
\newblock {\em Journal of Machine Learning Research}, 19(84):1--5.

\bibitem[\protect\astroncite{Biecek and Burzykowski}{2019}]{pmvee2019}
Biecek, P. and Burzykowski, T. (2019).
\newblock Explanatory model analysis. explore, explain and examine predictive
  models.
\newblock online.

\bibitem[\protect\astroncite{Bischl et~al.}{2014}]{bis2014onc}
Bischl, B., K{\"u}hn, T., and Szepannek, G. (2014).
\newblock On class imbalance correction for classification algorithms in credit
  scoring.
\newblock In Lübbecke, M., Koster, A., P., L., R., M., B., P., and Walther,
  G., editors, {\em Operations Research Proceedings}, pages 37--43.

\bibitem[\protect\astroncite{Brown and Christophe}{2012}]{brown2012}
Brown, I. and Christophe, M. (2012).
\newblock An experimental comparison of classification algorithms for
  imbalanced credit scoring data sets.
\newblock {\em Expert Systems with Applications}, 3(39):3446--3453.

\bibitem[\protect\astroncite{Bücker et~al.}{2013}]{buc2013rej}
Bücker, M., van Kampen, M., and Krämer, W. (2013).
\newblock Reject inference in consumer credit scoring with nonignorable missing
  data.
\newblock {\em Journal of Banking \& Finance}, 37(3):1040--1045.

\bibitem[\protect\astroncite{Chen et~al.}{}]{chenetal}
Chen, C., Lin, K., Rudin, C., Shaposhnik, Y., Wang, S., and Wang, T.
\newblock An explainable model for credit risk performance.
\newblock Explainable Machine Learning Challenge documentation.

\bibitem[\protect\astroncite{Chen and Guestrin}{2016}]{Chen}
Chen, T. and Guestrin, C. (2016).
\newblock Xgboost: A scalable tree boosting system.
\newblock In {\em Proceedings of the 22Nd ACM SIGKDD International Conference
  on Knowledge Discovery and Data Mining}, KDD '16, pages 785--794, New York,
  NY, USA. ACM.

\bibitem[\protect\astroncite{Cook}{2016}]{h2obook}
Cook, D. (2016).
\newblock {\em {Practical Machine Learning with H2O: Powerful, Scalable
  Techniques for Deep Learning and AI}}.
\newblock O'Reilly Media.

\bibitem[\protect\astroncite{Cortes and Vapnik}{1995}]{Cortes1995}
Cortes, C. and Vapnik, V. (1995).
\newblock Support-vector networks.
\newblock {\em Machine Learning}, 20(3):273--297.

\bibitem[\protect\astroncite{Crook et~al.}{2007}]{crook2007}
Crook, J., Edelman, D., and Thomas, L.~C. (2007).
\newblock Recent developments in consumer credit risk assessment.
\newblock {\em Journal of the Operational Research Society}, (183):1447--1465.

\bibitem[\protect\astroncite{Dash et~al.}{2018}]{10.5555/3327345.3327376}
Dash, S., G\"{u}nl\"{u}k, O., and Wei, D. (2018).
\newblock Boolean decision rules via column generation.
\newblock In {\em Proceedings of the 32nd International Conference on Neural
  Information Processing Systems}, NIPS’18, page 4660–4670, Red Hook, NY,
  USA. Curran Associates Inc.

\bibitem[\protect\astroncite{{EU Expert Group on AI}}{2019}]{EthicsEU2019}
{EU Expert Group on AI} (2019).
\newblock Ethics guidelines for trustworthy ai.
\newblock Online.

\bibitem[\protect\astroncite{{European Banking Authority}}{2017}]{eba2017}
{European Banking Authority} (2017).
\newblock Guidelines on pd estimation, lgd estimation and the treatment of
  defaulted exposures.
\newblock Online.

\bibitem[\protect\astroncite{{European Commission}}{2020}]{Excellence20}
{European Commission} (2020).
\newblock On artificial intelligence - a european approach to excellence and
  trust.
\newblock Online.

\bibitem[\protect\astroncite{European~Union}{2016}]{EUdataregulations2018}
European~Union, G. D. P.~R. (2016).
\newblock Regulation (eu) 2016/679 of the european parliament and of the
  council.

\bibitem[\protect\astroncite{FICO}{2019}]{fic2018xml}
FICO (2019).
\newblock xml challenge.
\newblock Online.

\bibitem[\protect\astroncite{{Financial Stability Board}}{2017}]{fsb2017}
{Financial Stability Board} (2017).
\newblock Artificial intelligence and machine learning in financial services --
  market developments and financial stability implications.
\newblock Online.

\bibitem[\protect\astroncite{Finlay}{2012}]{fin2012cre}
Finlay, S. (2012).
\newblock {\em Credit Scoring, Response Modelling and Insurance Rating}.
\newblock Palgarve MacMillan.

\bibitem[\protect\astroncite{Fisher et~al.}{2018}]{figher2018vip}
Fisher, A., Rudin, C., and Dominici, F. (2018).
\newblock Model class reliance: Variable importance measures for any machine
  learning model class, from the 'rashomon' perspective.
\newblock {\em Journal of Computational and Graphical Statistics}.

\bibitem[\protect\astroncite{Fitzpatrick and Mues}{2016}]{fitz2016}
Fitzpatrick, T. and Mues, C. (2016).
\newblock An empirical comparison of classification algorithms for mortgage
  default prediction: evidence from a distressed mortgage market.
\newblock {\em European Journal of Operational Research}, 2(249):427--439.

\bibitem[\protect\astroncite{Friedman et~al.}{2010}]{glmnet}
Friedman, J., Hastie, T., and Tibshirani, R. (2010).
\newblock Regularization paths for generalized linear models via coordinate
  descent.
\newblock {\em Journal of Statistical Software}, 33(1):1--22.

\bibitem[\protect\astroncite{Friedman}{2000}]{Friedman00greedyfunction}
Friedman, J.~H. (2000).
\newblock Greedy function approximation: A gradient boosting machine.
\newblock {\em Annals of Statistics}, 29:1189--1232.

\bibitem[\protect\astroncite{Garzcarek and Steuer}{2019}]{garz2019}
Garzcarek, U. and Steuer, D. (2019).
\newblock {\em Approaching Ethical Guidelines for Data Scientists}, pages
  151--169.
\newblock Springer International Publishing.

\bibitem[\protect\astroncite{Gill and Hall}{2018}]{oreilly2018}
Gill, N. and Hall, P. (2018).
\newblock An introduction to machine learning interpretability.
\newblock O'Reilly Media, Inc.

\bibitem[\protect\astroncite{Goldstein et~al.}{2015}]{goldstein2015peeking}
Goldstein, A., Kapelner, A., Bleich, J., and Pitkin, E. (2015).
\newblock Peeking inside the black box: Visualizing statistical learning with
  plots of individual conditional expectation.
\newblock {\em Journal of Computational and Graphical Statistics},
  24(1):44--65.

\bibitem[\protect\astroncite{Gomez et~al.}{2020}]{Gomez2020}
Gomez, O., Holter, S., Yuan, J., and Bertini, E. (2020).
\newblock Vice: Visual counterfactual explanations for machine learning models.
\newblock In {\em Proceedings of the 25th International Conference on
  Intelligent User Interfaces}, IUI ’20, page 531–535. ACM.

\bibitem[\protect\astroncite{Goodman and Flaxman}{2017}]{Goodman_Flaxman_2017}
Goodman, B. and Flaxman, S. (2017).
\newblock European union regulations on algorithmic decision-making and a
  “right to explanation”.
\newblock {\em AI Magazine}, 38(3):50--57.

\bibitem[\protect\astroncite{{Gosiewska} and
  {Biecek}}{2019}]{2019arXiv190311420G}
{Gosiewska}, A. and {Biecek}, P. (2019).
\newblock {Do Not Trust Additive Explanations}.
\newblock {\em arXiv e-prints}, page arXiv:1903.11420.

\bibitem[\protect\astroncite{Greenwell et~al.}{2019}]{gbm}
Greenwell, B., Boehmke, B., Cunningham, J., and Developers, G. (2019).
\newblock {\em gbm: Generalized Boosted Regression Models}.
\newblock R package version 2.1.5.

\bibitem[\protect\astroncite{Greenwell}{2017}]{greenwell207pdp}
Greenwell, B.~M. (2017).
\newblock pdp: An r package for constructing partial dependence plots.
\newblock {\em The R Journal}, 9(1):421--436.

\bibitem[\protect\astroncite{Hand}{2009}]{hand2009}
Hand, D. (2009).
\newblock Measuring classifier performance: a coherent alternative to the area
  under the roc curve.
\newblock {\em Machine Learning}, 77:103–--123.

\bibitem[\protect\astroncite{Harrell}{2015}]{harrell2015regression}
Harrell, F. (2015).
\newblock {\em Regression Modeling Strategies: With Applications to Linear
  Models, Logistic and Ordinal Regression, and Survival Analysis}.
\newblock Springer Series in Statistics. Springer International Publishing.

\bibitem[\protect\astroncite{Holter et~al.}{}]{holteretal}
Holter, S., Gomez, O., and Bertini, E.
\newblock Fico explainable machine learning challenge. creating visual
  explanations to black-box machine learning models.
\newblock Explainable Machine Learning Challenge documentation.

\bibitem[\protect\astroncite{Jenkins et~al.}{2019}]{interpret2019}
Jenkins, S., Nori, H., Koch, P., and Caruana, R. (2019).
\newblock Interpretml.

\bibitem[\protect\astroncite{Kusner and Loftus}{2020}]{kusner2020}
Kusner, M. and Loftus, J. (2020).
\newblock The long road to fairer algorithms.
\newblock {\em Nature}, 534:34--36.

\bibitem[\protect\astroncite{Lessmann et~al.}{2015}]{les15ben}
Lessmann, S., Baesens, B., Seow, H.-V., and Thomas, L. (2015).
\newblock Benchmarking state-of-the-art classification algorithms for credit
  scoring: An update of research.
\newblock {\em European Journal of Operational Research}, 247(1):124--136.

\bibitem[\protect\astroncite{Liaw and Wiener}{2002}]{randomForest}
Liaw, A. and Wiener, M. (2002).
\newblock Classification and regression by randomforest.
\newblock {\em R News}, 2(3):18--22.

\bibitem[\protect\astroncite{Louzada et~al.}{2016}]{lou16clas}
Louzada, F., Ara, A., and Fernandes, G. (2016).
\newblock Classification methods applied to credit scoring: A systematic review
  and overall comparison.
\newblock {\em Surveys in OR and Management Science}, 21(2):117--134.

\bibitem[\protect\astroncite{Luebke et~al.}{2020}]{lueb2020}
Luebke, K., Gehrke, M., Horst, J., and Szepannek, G. (2020).
\newblock Why we should teach causal inference: Examples in linear regression
  with simulated data.
\newblock {\em Journal of Statistics Education}.

\bibitem[\protect\astroncite{Lundberg and Lee}{2017}]{NIPS2017_7062}
Lundberg, S.~M. and Lee, S.-I. (2017).
\newblock A unified approach to interpreting model predictions.
\newblock In Guyon, I., Luxburg, U.~V., Bengio, S., Wallach, H., Fergus, R.,
  Vishwanathan, S., and Garnett, R., editors, {\em Advances in Neural
  Information Processing Systems 30}, pages 4765--4774. Curran Associates, Inc.

\bibitem[\protect\astroncite{McGough}{2018}]{airQuality}
McGough, M. (2018).
\newblock {How bad is Sacramento’s air, exactly? Google results appear at
  odds with reality, some say}.
\newblock
  \url{https://www.sacbee.com/news/california/fires/article216227775.html}.
\newblock Accessed: 2019-10-12.

\bibitem[\protect\astroncite{Molnar}{2019}]{mol2019int}
Molnar, C. (2019).
\newblock {\em Interpretable Machine Learning}.
\newblock \url{https://christophm.github.io/interpretable-ml-book/}.

\bibitem[\protect\astroncite{Molnar et~al.}{2018}]{molnar2018iml}
Molnar, C., Bischl, B., and Casalicchio, G. (2018).
\newblock {iml: An R package for Interpretable Machine Learning}.
\newblock {\em Journal of Open Source Software}, 3(26):786.

\bibitem[\protect\astroncite{Montavon et~al.}{2018}]{montavon2018dsp}
Montavon, G., Samek, W., and Müller, K.-R. (2018).
\newblock Methods for interpreting and understanding deep neural networks.
\newblock {\em Digital Signal Processing}, 73:1 -- 15.

\bibitem[\protect\astroncite{O'Neil}{2016}]{ONeil}
O'Neil, C. (2016).
\newblock {\em Weapons of Math Destruction: How Big Data Increases Inequality
  and Threatens Democracy}.
\newblock Crown Publishing Group, New York, NY, USA.

\bibitem[\protect\astroncite{Płoński}{2019}]{mljar}
Płoński, P. (2019).
\newblock {\em {mljar-supervised: The Automated Machine Learning - the new
  standard in ML. Machine Learning for Humans}}.

\bibitem[\protect\astroncite{Ribeiro et~al.}{2016}]{lime}
Ribeiro, M.~T., Singh, S., and Guestrin, C. (2016).
\newblock "why should {I} trust you?": Explaining the predictions of any
  classifier.
\newblock In {\em Proceedings of the 22nd {ACM} {SIGKDD} International
  Conference on Knowledge Discovery and Data Mining, San Francisco, CA, USA,
  August 13-17, 2016}, pages 1135--1144.

\bibitem[\protect\astroncite{Robin et~al.}{2011}]{robin2011}
Robin, X., Turck, N., Hainard, A., Tiberti, N., Lisacek, F., Sanchez, J., and
  Müller, M. (2011).
\newblock proc: an open-source package for r and s+ to analyze and compare roc
  curves.
\newblock {\em BMC Bioinformatics}, 12.

\bibitem[\protect\astroncite{Rudin}{2019}]{Rudin2019}
Rudin, C. (2019).
\newblock Stop explaining black box machine learning models for high stakes
  decisions and use interpretable models instead.
\newblock {\em Nature Machine Intelligence}, 1(5):206--215.

\bibitem[\protect\astroncite{Scallan}{2011}]{sca2011clas}
Scallan, G. (2011).
\newblock Class(ic) scorecards – selecting attributes in logistic regression.
\newblock In {\em Credit Scoring and Credit Control XIII}.

\bibitem[\protect\astroncite{Schölkopf}{2019}]{schoelkopf2019}
Schölkopf, B. (2019).
\newblock Causality for machine learning.

\bibitem[\protect\astroncite{Sokol and Flach}{2020}]{Sokol2020}
Sokol, K. and Flach, P. (2020).
\newblock {One Explanation Does Not Fit All}.
\newblock {\em KI - Kunstliche Intelligenz}.

\bibitem[\protect\astroncite{Szepannek}{2017a}]{sze2017afr}
Szepannek, G. (2017a).
\newblock A framework for scorecard modelling using r.
\newblock In {\em Credit Scoring and Credit Control XV}.

\bibitem[\protect\astroncite{Szepannek}{2017b}]{sze2017ont}
Szepannek, G. (2017b).
\newblock On the practical relevance of modern machine learning algorithms for
  credit scoring applications.
\newblock {\em WIAS Report Series}, 29:88--96.

\bibitem[\protect\astroncite{Szepannek}{2019}]{sze2019how}
Szepannek, G. (2019).
\newblock How much can we see? {A} note on quantifying explainability of
  machine learning models.
\newblock arxiv.

\bibitem[\protect\astroncite{Szepannek and Aschenbruck}{2019}]{sze2019pre}
Szepannek, G. and Aschenbruck, R. (2019).
\newblock Predicting ebay prices: Selecting and interpreting machine learning
  models – results of the ag dank 2018 data science competition.
\newblock {\em Archives of Data Science A (accepted)}.

\bibitem[\protect\astroncite{Thomas et~al.}{2019}]{tho2002cre}
Thomas, L.~C., Crook, J.~N., and Edelman, D.~B. (2019).
\newblock {\em Credit Scoring and its Applications}.
\newblock SIAM, second edition.

\bibitem[\protect\astroncite{Tobback and Martens}{2019}]{tobback2019}
Tobback, E. and Martens, D. (2019).
\newblock Retail credit scoring using fine-grained payment data.
\newblock {\em Journal of the Royal Statistical Society: Series A (Statistics
  in Society)}, 182(4):1227--1246.

\bibitem[\protect\astroncite{Verbraken et~al.}{2014}]{verbraken2014}
Verbraken, T., Bravo, C., Richard, W., and Baesens, B. (2014).
\newblock Development and application of consumer credit scoring models using
  profit-based classification measures.
\newblock {\em European Journal of Operational Research}, 238(2):505--513.

\bibitem[\protect\astroncite{Wexler}{2017}]{nytimesJail}
Wexler, R. (2017).
\newblock {When a Computer Program Keeps You in Jail}.
\newblock
  \url{https://www.nytimes.com/2017/06/13/opinion/how-computers-are-harming-criminal-justice.html}.
\newblock Accessed: 2019-10-12.

\bibitem[\protect\astroncite{Wright and Ziegler}{2017}]{ranger}
Wright, M.~N. and Ziegler, A. (2017).
\newblock {ranger}: A fast implementation of random forests for high
  dimensional data in {C++} and {R}.
\newblock {\em Journal of Statistical Software}, 77(1):1--17.

\bibitem[\protect\astroncite{Zhao and Hastie}{2019}]{zhao2019}
Zhao, Q. and Hastie, T. (2019).
\newblock Causal interpretations of black-box models.
\newblock {\em Journal of Business \& Economic Statistics}.

\end{thebibliography}

\end{document}